\documentclass{article}

 \usepackage[preprint]{neurips_2026}

\usepackage[utf8]{inputenc}
\usepackage[T1]{fontenc}
\usepackage{hyperref}
\usepackage{url}
\usepackage{booktabs}
\usepackage{amsfonts}
\usepackage{amsmath}
\usepackage{amssymb}
\usepackage{mathtools}
\usepackage{nicefrac}
\usepackage{microtype}
\usepackage{xcolor}
\usepackage{graphicx}
\usepackage{subcaption}
\usepackage{multirow}
\usepackage{array}
\usepackage{placeins}
\usepackage{enumitem}
\usepackage{placeins}
\usepackage{float}

\providecommand{\scoreerr}[1]{{\tiny $\pm$#1}}

\newcommand{\method}{ConRec}
\newcommand{\rbh}{RBH}

\newcommand{\eps}{\varepsilon}

\title{MANAS-2: Constrained Reconstruction for EEG Foundation Models}

\author{%
	Arvasu Kulkarni\thanks{These authors contributed equally.}\\
	Mannas AI\\
	\texttt{arvasu@mannas.ai}
	\And
	Aditya Ray Mishra\footnotemark[1]\\
	Mannas AI\\
	\texttt{aditya@mannas.ai}
	\And
	Jeet Bandhu Lahiri\\
	Indian Institute of Technology, Mandi\\
	\texttt{d23146@students.iitmandi.ac.in}
	\And
	Mahir Jain\\
	Mannas AI\\
	\texttt{mahir@mannas.ai}
	\And
	Parshva Runwal\\
	Mannas AI\\
	\texttt{parshva@mannas.ai}
	\And
	Lakshya Saini\\
	Mannas AI\\
	\texttt{lakshya@mannas.ai}
	\And
	Siddharth Panwar\\
	Mannas AI\\
	\texttt{siddharth@mannas.ai}
	\And
	Sandeep Singh\\
	Mannas AI\\
	\texttt{sandeep@mannas.ai}
}

\begin{document}
\maketitle
\setcounter{footnote}{0}

\begin{abstract}
Masked reconstruction is widely used for EEG foundation models, but optimizing reconstruction on low-SNR waveforms does not necessarily produce the most useful latent representation. We introduce \textsc{MANAS-2}, a new EEG foundation model that combines a Raw-Band Hybrid (\rbh{}) masked autoencoder with Constrained Reconstruction (\method{}), a physics--motivated regularizer. \rbh{} jointly reconstructs temporal waveform patches and compact spectral-band targets, while \method{} acts only on the temporal decoder output, penalizing differences in RMS energy between adjacent short windows of the reconstructed waveform. \method{} is intended to shape the encoder by biasing it toward the organization of oscillatory-envelope information. Across seven held-out EEG datasets, adding \method{} to an otherwise identical \rbh{} model increases frozen ridge recovery of six-band spectral power from mean $R^2=0.860$ to $0.906$ and recovery of inter-patch band-energy dynamics from $R^2=0.283$ to $0.354$, while temporal waveform information remains highly recoverable from the frozen latents. Applied to a temporal-only masked autoencoder, \method{} also improves frozen downstream transfer and frequency-dependent latent geometry despite receiving no spectral targets: i.e., the effects of \method{} are architecture-independent. \textsc{MANAS-2} also outperforms leading EEG Foundation Models on most downstream knowledge-transfer tasks. From the effects of \method{}, we see that a physically motivated constraint imposed through the decoder can make for a more spectrally organized and transferable latent space. \textsc{MANAS-2} therefore provides a new EEG foundation model built around constrained reconstruction as a mechanism for shaping representation -- rather than reconstruction -- quality.\end{abstract}

\section{Introduction}

Self-supervised pretraining has become a standard route to general-purpose EEG encoders. Contrastive methods, neural tokenizers, and masked autoencoders have shown that pretrained EEG representations can transfer across subjects, tasks, and recording conditions~\citep{kostas2021bendr,yang2023biot,jiang2024lbm,wang2024eegpt,jiang2025neurolm,yuan2024brainwave,kuruppu2026eeg}. Yet the objective used by many EEG foundation models remains close to the default objective used for generic sequences: mask patches and reconstruct their raw values. This design is simple, but it treats the signal as a bag of local fragments rather than as a continuous physical measurement.

EEG differs from generic time series in three ways that matter for pretraining. First, it is low-SNR: optimizing only pointwise waveform error can reward reconstruction of nuisance variation. Second, EEG is oscillatory: clinically and cognitively meaningful structure is often expressed through canonical frequency bands and their evolving energy envelopes~\citep{newson2019eeg,cao2022brain}. Third, EEG is continuous: adjacent patches are not independent samples but neighboring segments of an electrical potential. A masked reconstruction objective that treats these patches independently does not impose any prior on how local signal statistics should evolve across the reconstructed waveform.

This paper studies whether such a prior can be used to shape the learned representation rather than simply improve waveform reconstruction. We introduce Constrained Reconstruction (\method), a temporal regularizer for masked EEG pretraining. \method{} penalizes short-window RMS-energy jumps along regular temporal boundaries in the reconstructed waveform. Importantly, the \method{} regularizer itself has no spectral target. It acts only on the reconstructed time-domain waveform, using a physically motivated local energy constraint to bias \emph{what the encoder learns}.

Our proposed foundation model, \textsc{MANAS-2}, uses a novel Raw-Band Hybrid (\rbh{}) masked autoencoder architecture together with \method{}. The encoder receives temporal waveform patches; the primary decoder reconstructs temporal patches; and a dedicated band decoder predicts compact spectral-band targets aligned to the same patch grid. The necessity for this specific architecture---with comparisons against a temporal MAE, naive spectral loss, pure spectral tokenization, hybrid variants, and vanilla \rbh{}---is reported in Appendix~\ref{app:objective_family}.

Our central result is that this reconstruction-space regularization reshapes spectral representations even though it does not explicitly impact - or even require - spectral decoding. Across seven held-out EEG datasets, \textsc{MANAS-2} improves frozen ridge recovery of six-band spectral power from mean $R^2=0.860$ for \rbh{} to $0.906$ and improves recovery of adjacent-patch band-energy dynamics from $R^2=0.283$ to $0.354$, while preserving high temporal waveform recoverability from the latent space. Applying \method{} to a temporal-only masked autoencoder also improves downstream transfer and spectral latent geometry; these recoverability gains are strongest when \method{} is paired with the \rbh{} architecture. These findings support a constraints-vs-targets principle for EEG pretraining: the target specifies what the decoder is asked to reconstruct, while auxiliary constraints on that reconstruction can bias how that information is organized in the encoder.

\paragraph{Contributions.}
We make three contributions. First, we define \method{}, a temporal constrained-reconstruction regularizer for EEG masked modeling. Second, we show that \method{} improves recoverability of spectral power and band-energy dynamics in hybrid architectures, and alters spectral latent geometry across architectures despite acting only on reconstructed temporal waveforms. Third, we propose \textsc{MANAS-2}, an EEG foundation model built on a novel spectral-temporal Raw-band Hybrid (\rbh{}) architecture together with \method{}. \textsc{MANAS-2} produces better downstream performance across seven held-out datasets than most leading EEG foundation models in the literature.

\section{Related work}
\label{sec:related}
This section summarises recent work related to our explorations.

\paragraph{EEG foundation models.}
Self-supervised EEG pretraining has evolved from contrastive learning~\citep{kostas2021bendr} to masked autoencoding and tokenization-based approaches~\citep{yang2023biot,jiang2024lbm,wang2024eegpt,jiang2025neurolm,yuan2024brainwave}. Recent surveys emphasize both the promise of cross-dataset transfer and the instability of evaluation protocols for EEG foundation models~\citep{kuruppu2026eeg}. Our work is complementary: rather than proposing a larger encoder, we study how reconstruction constraints shape the information stored in frozen latents.

\paragraph{Spectral and modality-aware objectives.}
EEG contains strong spectral semantics, motivating objectives that reconstruct or otherwise model time--frequency structure~\citep{jiang2024lbm,shi2026fome,guo2026brainof,darankoum2026specmoe}. Our results show that explicit spectral targets are not the only factor shaping spectral representations. Temporal energy-plausibility constraints can encourage preservation of spectral-envelope information because local energy imposes broad spectral pressure.

\paragraph{Representation probing and constrained learning.}
Frozen probes are commonly used to diagnose what is linearly available in intermediate representations~\citep{alain2017understanding}. Our probes extend this idea to EEG-specific targets: six-band power, temporal waveform recovery, and bandflow. Conceptually, \method{} is related to constrained and physics-informed learning~\citep{raissi2019pinn}: instead of only matching targets, the model must satisfy structural conditions that valid signals obey.

\section{Method}
\label{sec:method}

\textsc{MANAS-2} uses a masked-autoencoding scaffold~\citep{he2022masked}, adapted to EEG through REVE-style temporal-patch tokenization, spatiotemporal positional encoding, block masking, and a global auxiliary reconstruction head~\citep{elouahidi2025reve}. EEG is resampled to 200\,Hz and represented as overlapping temporal patch tokens over the channel--time grid. On top of this, \textsc{MANAS-2} introduces two components: a novel Raw-Band Hybrid (\rbh{}) architecture and Constrained Reconstruction (\method{}). \rbh{} introduces explicit spectral supervision through a dedicated decoder while retaining temporal waveform tokens, whereas \method{} applies a physically motivated local energy prior through the temporal reconstruction pathway to shape the shared encoder representation.

The complete pretraining objective is
\begin{equation}
	\mathcal{L}
	=
	\mathcal{L}_{\mathrm{pri}}
	+
	\lambda_{\mathrm{sec}}\mathcal{L}_{\mathrm{sec}}
	+
	\lambda_{\mathrm{band}}\mathcal{L}_{\mathrm{band}}
	+
	\lambda_r\mathcal{L}_{\mathrm{rms}},
\end{equation}
where the first two terms reconstruct temporal waveform targets, $\mathcal{L}_{\mathrm{band}}$ is the \rbh{} spectral objective, and $\mathcal{L}_{\mathrm{rms}}$ is the \method{} regularizer.

\begin{figure*}[!htbp]
	\centering
	\includegraphics[width=\textwidth]{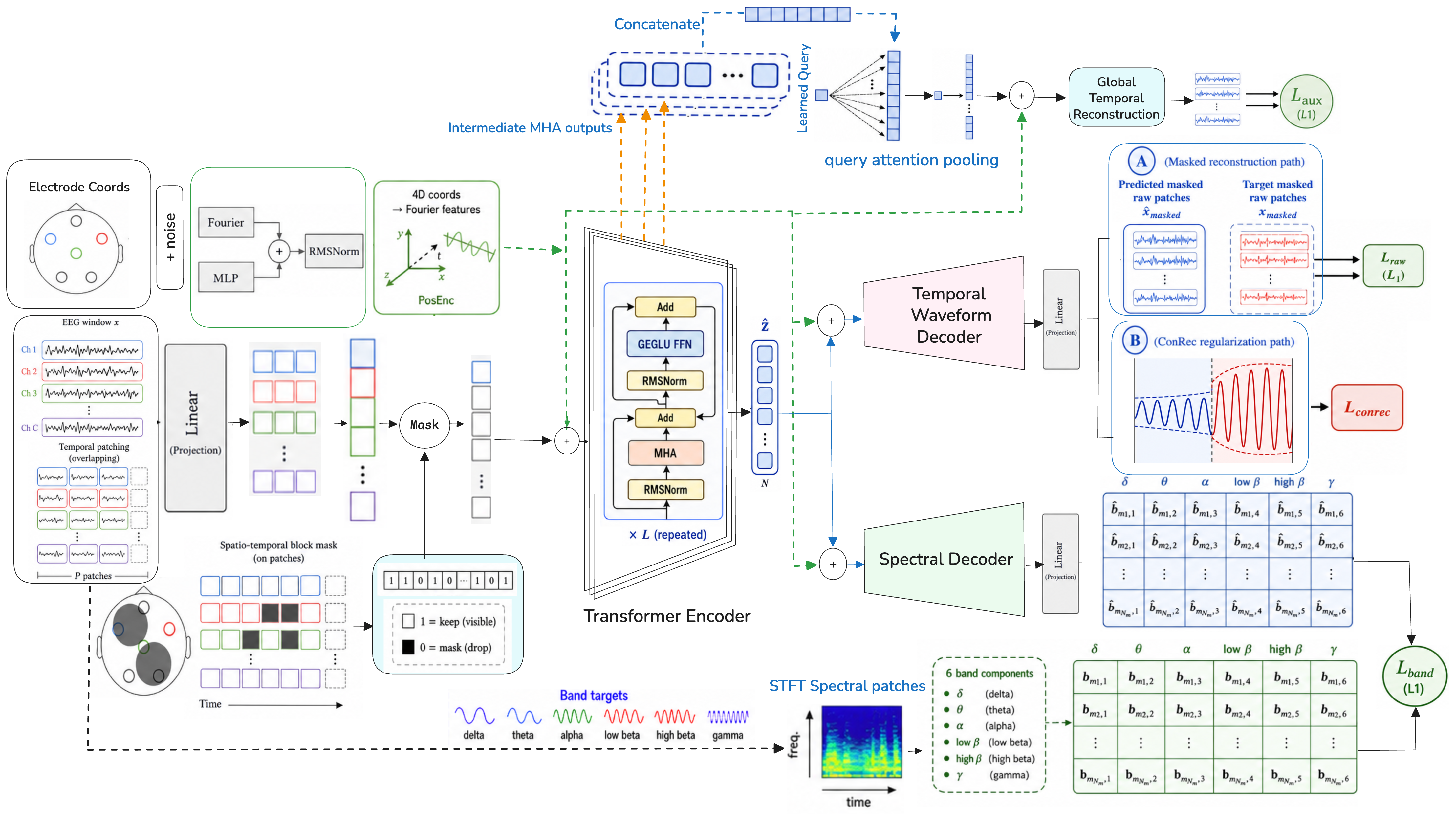}
	\caption{
		\textbf{MANAS-2 architecture.}
		An EEG signal is represented as overlapping temporal patch tokens with spatiotemporal positional encodings. The tokens are processed by a transformer encoder, followed by temporal and spectral decoding. The \method{} regularizer is applied to the temporal decoder output.
	}
	\label{fig:conrec_rbh_architecture}
\end{figure*}

\paragraph{Raw-Band Hybrid.}
\rbh{} retains temporal waveform patches as the encoder input and primary reconstruction target, but adds a dedicated spectral decoder operating on the same visible encoder tokens. For each temporal patch, a frozen STFT frontend constructs normalized log-power targets over six canonical EEG bands: $\delta$ (0.5--4\,Hz), $\theta$ (4--8\,Hz), $\alpha$ (8--13\,Hz), low-$\beta$ (13--20\,Hz), high-$\beta$ (20--30\,Hz), and $\gamma$ (30--40\,Hz). The spectral decoder is trained with an L1 loss on masked tokens. Comparisons against temporal-only reconstruction, direct STFT supervision, and spectral-token alternatives are given in Appendix~\ref{app:objective_family}.

\paragraph{Constrained Reconstruction.}
\method{} acts on the reconstructed time-domain waveform $\hat{\mathbf{x}}$ produced by the temporal decoder. Regularization boundaries are placed every $\ell=200$ samples (1\,s). At each valid boundary $\tau$, \method{} compares RMS energy in windows of length $q$ immediately before and after the boundary:
\begin{equation}
	r^{\mathrm{rms}}_{b,c,\tau}
	=
	\left(
	\frac{1}{q}
	\sum_{u=\tau-q}^{\tau-1}
	\hat{x}_{b,c,u}^{2}
	+ \eps
	\right)^{1/2}
	-
	\left(
	\frac{1}{q}
	\sum_{u=\tau}^{\tau+q-1}
	\hat{x}_{b,c,u}^{2}
	+ \eps
	\right)^{1/2}.
\end{equation}
$\mathcal{L}_{\mathrm{rms}}$ applies a Smooth-L1 penalty to these residuals and averages over batches, channels, and valid boundaries. In \textsc{MANAS-2}, $q=32$, $\eps=10^{-6}$, and $\lambda_r=3.0$. Unlike the spectral branch, \method{} introduces no additional prediction target; it uses a local energy prior through the decoder pathway to reshape the shared encoder representation.

\section{Experiments}
\label{sec:experiments}
\subsection{Setup and probes}

We evaluate frozen representations on seven held-out EEG datasets through (i) latent-space probes and (ii) frozen downstream transfer; no encoder parameters are updated. The latent probes measure six-band power, adjacent-patch changes in band power (\emph{bandflow}), Peak-Alpha Frequency (PAF), the periodic component and aperiodic exponent obtained using FOOOF~\citep{donoghue2020fooof}, and time-domain waveform recoverability.  Because both six-band power and bandflow are closely related to the \rbh{} spectral target, PAF, periodic structure, and aperiodic exponent provide complementary spectral properties not directly supervised during pretraining. Waveform recoverability serves as a control for whether spectral gains are accompanied by loss of accessible time-domain information.

\subsection{Spectral information recoverability}
\label{subsec:specbandflow}

\begin{table*}[!htbp]
	\caption{
		Frozen ridge-probe recovery across seven held-out EEG datasets. Values are mean test $R^2$ across datasets. Full dataset-wise results are reported in Appendix~\ref{app:stftbandflow_details}.
	}
	\label{tab:latent_probes}
	\centering
	\small
	\setlength{\tabcolsep}{6pt}
	\begin{tabular}{@{}lcccccc@{}}
		\toprule
		Model
		& Six-band
		& Bandflow
		& PAF
		& Periodic
		& Aperiodic
		& Waveform \\
		\midrule
		Temporal MAE
		& .747
		& .165
		& .015
		& .067
		& .678
		& \textbf{.927} \\
		
		\textsc{RBH}
		& .860
		& .283
		& .135
		& .203
		& .788
		& .920 \\
		
		\textsc{MANAS-2}
		& \textbf{.906}
		& \textbf{.354}
		& \textbf{.208}
		& \textbf{.254}
		& .792
		& .917 \\
		\bottomrule
	\end{tabular}
\end{table*}

Table~\ref{tab:latent_probes} contains the central empirical result. Relative to \textsc{RBH}, \textsc{MANAS-2} substantially improves the spectral information recoverable from frozen latents even though the only additional objective is \method{}, which acts on the reconstructed temporal waveform and introduces no new spectral target. Mean six-band recovery increases from $R^2=0.860$ to $0.906$, while bandflow increases from $0.283$ to $0.354$, showing a particularly strong gain in the representation of spectral-energy dynamics.

The effect extends beyond the spectral quantities directly supervised by the \rbh{} decoder. PAF recovery increases from $R^2=0.135$ to $0.208$, and recovery of the periodic PSD component from $0.203$ to $0.254$. In contrast, the aperiodic exponent changes little ($0.788$ to $0.792$), consistent with \method{} acting primarily on local energy and oscillatory-envelope organization rather than broadband spectral slope. Waveform recoverability remains high ($0.920$ to $0.917$), indicating that these spectral gains do not arise from substantially discarding accessible time-domain information.

\subsection{Latent geometry}

\label{sec:synthetic_geometry}

We next test whether \method{} changes the organization of spectral information in the latent space, rather than only its recoverability. We generate artificial signals: fixed-frequency sinusoids spanning 0.5--40\,Hz in 0.5\,Hz increments, with five repetitions per frequency, each with a different sample of low-amplitude noise added. We evaluate both single-channel (Cz) inputs and 18-channel inputs with independently randomized phase across channels. Because frequency is constant within each signal, this diagnostic isolates frequency representation from evolving spectral dynamics.

For each artificial sinusoid, we average the encoder’s output vectors over time and channels to obtain a single representation. We then average these representations across the five repetitions at each frequency, yielding one representative vector per frequency, which we call its \textit{latent centroid}. We then measure whether physical frequency separation is reflected in latent distance by computing the Spearman correlation between $|f_i-f_j|$ and the Euclidean distance between the corresponding frequency centroids. Higher $\rho_f$ indicates stronger ordering of the latent space by physical frequency.

\begin{table*}[!htbp]
	
	\caption{Frequency-distance ordering in the latent space, measured by Spearman $\rho_f$ (higher is better).}
	
	\label{tab:native_euclidean_frequency_geometry}
	
	\centering
	
	\small
	
	\setlength{\tabcolsep}{8pt}
	
	\begin{tabular}{@{}lcccc@{}}
		
		\toprule
		
		Condition & Temporal MAE & \method{}-Temporal MAE& \textsc{RBH} & \textsc{MANAS-2} \\
		
		\midrule
		
		Single Cz
		
		& 0.816 & \textbf{0.857} & 0.792 & \textbf{0.822} \\
		
		18-channel random phase
		
		& 0.765 & \textbf{0.820} & 0.767 & \textbf{0.860} \\
		
		\bottomrule
		
	\end{tabular}
	
\end{table*}

Table~\ref{tab:native_euclidean_frequency_geometry} shows that \method{} improves frequency-distance ordering in both architectural families and under both input conditions. Within the \rbh{} family, \textsc{MANAS-2} increases $\rho_f$ from $0.792$ to $0.822$ for single-channel inputs and from $0.767$ to $0.860$ for the 18-channel condition. The same effect in the temporal-only ablation shows that \method{} changes spectral latent organization even without an explicit spectral decoder, while the strongest ordering is observed in \textsc{MANAS-2}.

\FloatBarrier

\subsection{Frozen downstream transfer}
\label{sec:downstream_transfer}

We evaluate frozen transfer using average pooling over encoder tokens and a lightweight classifier head (\textsc{LP-Avg}), with no encoder updates. Table~\ref{tab:frozen_downstream_transfer} compares \textsc{MANAS-2} against six leading EEG foundation models under the same protocol.

\providecommand{\scoreerr}[1]{{\tiny $\pm$#1}}
\begin{table*}[!htbp]
\caption{
Primary frozen downstream transfer: \textsc{LP-Avg} probe with average pooling and a lightweight classifier head.
Each model has subrows for balanced accuracy, F1/AUROC, and $\kappa$/AUC-PR.
Values are mean test score (\%) at the best validation epoch, $\pm$ reported variation across seeds.
}
\label{tab:frozen_downstream_transfer}
\centering
\renewcommand{\arraystretch}{0.90}
\setlength{\aboverulesep}{0.3ex}
\setlength{\belowrulesep}{0.45ex}
\scriptsize
\setlength{\tabcolsep}{2pt}
\resizebox{\textwidth}{!}{%
\begin{tabular}{@{}llccccccc@{}}
\toprule
Model & Metric & ADFTD & BCIC-2a & HMC & Motor$^{\ast}$ & Siena & Workload & MIMUL-11 \\
\midrule
EEGPT 
& Bal. Acc. & 39.1\scoreerr{1.5} & 26.6\scoreerr{1.3} & 65.5\scoreerr{0.9} & 38.7\scoreerr{1.5} & \textbf{81.0}\scoreerr{3.5} & 68.4\scoreerr{2.0} & 38.8\scoreerr{1.1} \\
& F1/AUROC & 39.9\scoreerr{2.5} & 16.7\scoreerr{1.6} & 70.2\scoreerr{0.9} & 38.2\scoreerr{1.8} & \textbf{96.0}\scoreerr{0.5} & 74.6\scoreerr{0.4} & 43.6\scoreerr{3.9} \\
& $\kappa$/AUC-PR & 12.2\scoreerr{2.9} & 2.1\scoreerr{1.8} & 61.9\scoreerr{1.1} & 18.2\scoreerr{2.0} & \textbf{76.6}\scoreerr{4.5} & 45.5\scoreerr{1.2} & 9.6\scoreerr{2.1} \\
\midrule
CSBrain 
& Bal. Acc. & 40.8\scoreerr{1.2} & 27.1\scoreerr{1.1} & 56.6\scoreerr{1.1} & 25.3\scoreerr{1.0} & 50.0\scoreerr{0.0} & 50.4\scoreerr{0.0} & 37.4\scoreerr{0.1} \\
& F1/AUROC & 42.8\scoreerr{1.0} & 15.5\scoreerr{2.3} & 59.9\scoreerr{1.0} & 24.4\scoreerr{1.3} & 19.6\scoreerr{6.7} & 54.1\scoreerr{0.8} & 44.3\scoreerr{0.5} \\
& $\kappa$/AUC-PR & 11.4\scoreerr{2.2} & 2.7\scoreerr{1.5} & 49.6\scoreerr{1.3} & 0.4\scoreerr{1.4} & 0.8\scoreerr{0.1} & 32.0\scoreerr{1.1} & 8.2\scoreerr{0.1} \\
\midrule
CBraMod 
& Bal. Acc. & 33.4\scoreerr{0.0} & 26.8\scoreerr{0.8} & 42.1\scoreerr{0.2} & 25.6\scoreerr{0.5} & 50.0\scoreerr{0.0} & 50.2\scoreerr{1.1} & 33.5\scoreerr{0.0} \\
& F1/AUROC & 22.9\scoreerr{0.0} & 20.0\scoreerr{2.4} & 45.9\scoreerr{0.3} & 19.6\scoreerr{1.4} & 9.4\scoreerr{0.1} & 51.5\scoreerr{4.5} & 40.2\scoreerr{0.2} \\
& $\kappa$/AUC-PR & 0.2\scoreerr{0.0} & 2.4\scoreerr{1.0} & 32.9\scoreerr{0.3} & 0.8\scoreerr{0.7} & 3.1\scoreerr{0.0} & 30.6\scoreerr{3.8} & 0.8\scoreerr{0.2} \\
\midrule
BIOT 
& Bal. Acc. & 45.1\scoreerr{3.3} & 26.6\scoreerr{2.7} & 64.0\scoreerr{0.6} & 26.8\scoreerr{1.0} & 67.4\scoreerr{1.2} & 55.5\scoreerr{5.6} & 38.1\scoreerr{2.4} \\
& F1/AUROC & 48.2\scoreerr{3.2} & 17.3\scoreerr{2.2} & 68.7\scoreerr{0.6} & 25.2\scoreerr{2.2} & 78.2\scoreerr{4.5} & 56.3\scoreerr{4.0} & 42.5\scoreerr{2.3} \\
& $\kappa$/AUC-PR & 20.1\scoreerr{5.3} & 2.2\scoreerr{3.6} & 59.3\scoreerr{0.8} & 2.4\scoreerr{1.3} & 38.7\scoreerr{4.7} & 37.3\scoreerr{6.9} & 7.4\scoreerr{3.5} \\
\midrule
REVE 
& Bal. Acc. & 43.2\scoreerr{2.9} & 27.1\scoreerr{0.4} & 61.8\scoreerr{0.5} & 27.4\scoreerr{0.9} & 67.4\scoreerr{2.7} & 69.7\scoreerr{2.2} & 36.5\scoreerr{0.9} \\
& F1/AUROC & 46.0\scoreerr{3.4} & 15.3\scoreerr{0.4} & 64.4\scoreerr{1.4} & 26.5\scoreerr{1.0} & 76.7\scoreerr{5.0} & 71.3\scoreerr{1.6} & 43.8\scoreerr{2.3} \\
& $\kappa$/AUC-PR & 18.3\scoreerr{5.5} & 2.8\scoreerr{0.5} & 55.3\scoreerr{1.5} & 3.2\scoreerr{1.2} & 31.6\scoreerr{6.5} & 51.3\scoreerr{2.7} & 5.3\scoreerr{2.2} \\
\midrule
LaBraM 
& Bal. Acc. & 30.9\scoreerr{2.1} & 24.9\scoreerr{1.3} & 34.9\scoreerr{0.4} & 25.6\scoreerr{0.5} & 50.0\scoreerr{0.0} & 49.9\scoreerr{0.3} & 33.7\scoreerr{0.2} \\
& F1/AUROC & 33.0\scoreerr{1.8} & 19.3\scoreerr{4.1} & 39.7\scoreerr{0.7} & 20.5\scoreerr{2.9} & 46.4\scoreerr{4.1} & 49.1\scoreerr{0.2} & 40.2\scoreerr{0.6} \\
& $\kappa$/AUC-PR & -3.0\scoreerr{3.6} & -0.1\scoreerr{1.7} & 23.6\scoreerr{0.5} & 0.8\scoreerr{0.7} & 1.1\scoreerr{0.1} & 27.5\scoreerr{0.2} & 0.8\scoreerr{0.4} \\
\midrule
\textbf{Ours: MANAS-2}
& Bal. Acc. & \textbf{47.7}\scoreerr{2.6} & \textbf{29.2}\scoreerr{0.5} & \textbf{70.3}\scoreerr{0.3} & \textbf{37.7}\scoreerr{0.7} & 80.0\scoreerr{1.2} & \textbf{77.8}\scoreerr{0.8} & \textbf{39.1}\scoreerr{0.4} \\
& F1/AUROC & \textbf{49.3}\scoreerr{3.9} & \textbf{20.5}\scoreerr{0.9} & \textbf{74.5}\scoreerr{0.4} & \textbf{37.2}\scoreerr{0.8} & 92.9\scoreerr{0.1} & \textbf{85.0}\scoreerr{0.7} & \textbf{46.1}\scoreerr{0.4} \\
& $\kappa$/AUC-PR & \textbf{27.3}\scoreerr{4.8} & \textbf{5.6}\scoreerr{0.7} & \textbf{66.9}\scoreerr{0.2} & \textbf{17.0}\scoreerr{0.9} & 73.4\scoreerr{0.5} & \textbf{58.8}\scoreerr{2.0} & \textbf{9.9}\scoreerr{0.5} \\
\bottomrule
\end{tabular}%
}
\vspace{2pt}
\begin{flushleft}
\scriptsize
$^{\ast}$Motor-MV was part of EEGPT's pretraining corpus; EEGPT's Motor results should therefore be interpreted with this overlap.
\end{flushleft}
\end{table*}

Overall, \textsc{MANAS-2} gives the best balanced accuracy on most downstream datasets. \footnote{The only exception is Motor-MV, where EEGPT is slightly higher and also has pretraining-corpus overlap, so that comparison should be interpreted cautiously. }
Together with the latent spectral and bandflow probes, these downstream results support the claim that temporal reconstruction constraints produce representations that are both more frequency-aware and more transferable.
\FloatBarrier

\section{Conclusion}

We introduced \textsc{MANAS-2}, an EEG foundation model that combines the Raw-Band Hybrid (\rbh{}) architecture with Constrained Reconstruction (\method{}), a physically motivated reconstruction-space regularizer for masked EEG pretraining. Using \rbh{} to jointly learn temporal waveform and compact spectral targets, we showed that \method{} improves knowledge transfer, spectral-power recovery, band-energy dynamics, and synthetic spectral latent geometry while preserving high waveform recoverability from the latent space. The key finding is that spectral organization can be strengthened by a local energy constraint applied through the temporal reconstruction pathway, not only by explicit spectral reconstruction targets. More broadly, these results suggest that EEG pretraining can be shaped jointly through reconstruction targets, architectural pathways, and reconstruction-space constraints, with \method{} biasing the encoder toward stronger organization of oscillatory-envelope information useful for downstream representation learning.

\appendix
\FloatBarrier

\section{Dataset description}
\label{app:dataset_description}

This section summarizes the EEG data used for self-supervised pretraining and downstream evaluation. 
Pretraining uses unlabeled EEG only. 
Downstream evaluation follows the EEG-FM-Bench task definitions and splits~\citep{xiong2026eegfmbench}; the pretrained encoder is frozen and only lightweight probe or classifier heads are trained.

\subsection{Pretraining datasets}
\label{app:pretraining_datasets}

We pretrain on large-scale unlabeled EEG recordings aggregated from multiple sources to capture diverse clinical and physiological variability. 
The pretraining corpus includes two internal clinical EEG sources, denoted \emph{Internal-A} and \emph{Internal-B}, together with public clinical EEG corpora. 
No diagnosis labels, event labels, sleep-stage labels, seizure labels, or downstream task labels are used in the self-supervised pretraining objective.

\paragraph{Internal clinical EEG corpus.}
Our internal clinical EEG corpus combines Internal-A and Internal-B. 
Internal-A contains 4,539 subjects and 2,546.19 hours of EEG, while Internal-B contains 1,050 subjects and 435.43 hours. 
Together, these internal sources contain 5,589 subjects and 2,981.62 hours of EEG. 
Both internal sources are processed with the same resampling, filtering, coordinate mapping, windowing, and sharding pipeline before being mixed with the public pretraining corpora.

\paragraph{TUH EEG.}
The Temple University Hospital EEG Corpus contains more than 15000 subjects with approximately 25,000 hours of multichannel EEG recordings collected in clinical settings~\citep{obeid2016tuh}. 
We use TUH EEG as a large, clinically diverse source of unlabeled EEG for masked reconstruction pretraining.

\paragraph{I-CARE.}
The International Cardiac Arrest REsearch consortium database contains EEG recordings from 600 patients after cardiac arrest, with continuous monitoring over 33,000 hours across intensive-care settings~\citep{amorim2023icare}. 
We include I-CARE to expose the model to long-duration critical-care EEG and clinically relevant changes in background rhythm and temporal dynamics.

\begin{table*}[!htbp]
\caption{
Self-supervised pretraining sources. 
Internal-A and Internal-B are combined as our internal clinical EEG corpus. 
The internal corpus counts reflect the data included in our pretraining pool. 
TUH EEG and I-CARE values are corpus-level reference descriptions from the cited sources; effective training hours may differ after preprocessing, channel filtering, and quality-control exclusions.
}
\label{tab:pretraining_sources}
\centering
\tiny
\setlength{\tabcolsep}{2pt}
\resizebox{0.85\textwidth}{!}{%
\begin{tabular}{@{}lccc@{}}
\toprule
Source & Subjects / patients & EEG duration & Role in pretraining \\
\midrule
Internal-A & 4,539 & 2,546 hours & Internal clinical EEG \\
Internal-B & 1,050 & 435 hours & Internal clinical EEG \\
\cmidrule(lr){1-4}
{Internal total} & {5,589} & {2,981 hours} & Internal clinical variability \\
\midrule
TUH EEG~\citep{obeid2016tuh} & $>15{,}000$ & $\sim$25,000 hours & Public clinical EEG \\
I-CARE~\citep{amorim2023icare} & 600 & $\sim$33,000 hours & Critical-care EEG \\
\bottomrule
\end{tabular}%
}
\end{table*}

\subsection{Downstream datasets}
\label{app:downstream_datasets}

We evaluate pretrained representations on seven datasets from EEG-FM-Bench~\citep{xiong2026eegfmbench}, spanning clinical, motor, sleep, seizure, cognitive, and upper-extremity movement tasks. 
The suite is heterogeneous in subject count, recording duration, montage size, window length, and class imbalance, which makes it useful for testing whether learned representations transfer beyond the pretraining objective.

\paragraph{ADFTD.}
ADFTD is a resting-state clinical EEG dataset for distinguishing Alzheimer's disease, frontotemporal dementia, and cognitively normal controls. 
The dataset contains 88 subjects recorded with a 19-channel clinical EEG montage~\citep{miltiadous2023adftd}. 
We use it as a three-class clinical classification task.

\paragraph{BCIC-IV 2A.}
BCI Competition IV dataset 2a is a cue-based motor-imagery benchmark with 9 subjects, 22 EEG channels, and four imagined movement classes: left hand, right hand, feet, and tongue~\citep{tangermann2012bci}. 
We use it as a four-class motor imagery task.

\paragraph{HMC.}
The Haaglanden Medisch Centrum sleep staging database is a whole-night polysomnography corpus with EEG and other physiological channels, scored in 30\,s epochs into wake, N1, N2, N3, and REM sleep stages~\citep{alvarez2022hmc,alvarez2021interdatabase}. 
We use the EEG channels provided by the benchmark as a five-class sleep staging task.

\paragraph{PhysioNet MI.}
PhysioNet MI uses the EEG Motor Movement/Imagery dataset, which contains motor execution and motor imagery recordings from 109 volunteers with 64-channel EEG acquired using BCI2000~\citep{physionet_eegmmidb,schalk2004bci2000}. 
We use the EEG-FM-Bench four-class motor imagery formulation.

\paragraph{Siena.}
The Siena scalp EEG dataset contains long-term scalp EEG recordings from 14 epilepsy patients, with expert-annotated seizure events~\citep{detti2020siena}. 
We use it as a binary seizure detection task.

\paragraph{EEGMAT.}
EEGMAT is derived from the EEG During Mental Arithmetic Tasks dataset, which records EEG before and during serial-subtraction mental arithmetic~\citep{zyma2019eegmat}. 
We use it as a binary workload classification task.

\paragraph{MIMUL-11.}
MIMUL-11 is based on a multimodal upper-extremity movement dataset containing EEG, EMG, and EOG from 25 healthy participants performing intuitive arm and hand movement tasks across multiple recording sessions~\citep{jeong2020mimul}.

\begin{table*}[!htbp]
\caption{
Downstream dataset statistics. 
All datasets are used through the EEG-FM-Bench downstream protocol~\citep{xiong2026eegfmbench}. 
}
\label{tab:downstream_dataset_stats}
\centering
\tiny
\setlength{\tabcolsep}{2pt}
\resizebox{0.90\textwidth}{!}{%
\begin{tabular}{@{}lcccll@{}}
\toprule
Dataset & Subjects & Duration (hrs) & EEG channels & Classes & Task \\
\midrule
ADFTD & 88 & $\sim$19.4 & 19 & 3 & Clinical classification \\
BCIC-IV 2A & 9 & $\sim$6 & 22 & 4 & Motor imagery \\
HMC & 151 & $\sim$1200 & 4 & 5 & Sleep staging \\
PhysioNet MI & 109 & $\sim$50 & 64 & 4 & Motor imagery \\
Siena & 14 & $\sim$128 & 29 & 2 & Seizure detection \\
EEGMAT & 36 & $\sim$2.4 & 21 & 2 & Workload classification \\
MIMUL-11 & 25 & $\sim$54.2 & 60 & 3 & Upper-extremity movement classification \\
\bottomrule
\end{tabular}%
}
\end{table*}

\subsection{Preprocessing}
\label{app:preprocessing}

Raw EEG files are converted into a common pretraining and evaluation format. 
Signals are resampled, filtered, windowed into fixed-length segments, mapped to electrode-coordinate metadata when channel labels are available, and stored as training samples. 
All downstream evaluations use the EEG-FM-Bench splits and label definitions~\citep{xiong2026eegfmbench}; any dataset-specific channel mapping follows the benchmark protocol.

\FloatBarrier

\section{Additional Experimental Details}
\label{app:details}

Unless otherwise stated, all models and objective ablations use the same self-supervised pretraining setup. The EEG signals are resampled to 200\,Hz and segmented into 10\,s windows. Key architectural and optimization settings are summarized below.

\begin{table*}[!htbp]
	\caption{Key pretraining configuration for \textsc{MANAS-2} and its ablations.}
	\label{tab:pretrain_config}
	\centering
	\renewcommand{\arraystretch}{0.86}
	\small
	\setlength{\tabcolsep}{7pt}
	\begin{tabular}{@{}ll@{}}
		\toprule
		Setting & Value \\
		\midrule
		\multicolumn{2}{@{}l}{\textbf{Architecture and training}} \\
		Encoder & 22 Transformer layers, $d=512$, 8 heads \\
		Decoder & 4 Transformer layers, 8 heads \\
		Input window & 10\,s at 200\,Hz \\
		Temporal patches & 1.0\,s, 0.1\,s overlap \\
		Masking ratio & 0.55 \\
		Optimizer & AdamW, lr $2.4\times10^{-4}$, weight decay 0.05 \\
		Effective batch size & 2048 \\
		Precision & bfloat16 mixed precision \\
		\midrule
		\multicolumn{2}{@{}l}{\textbf{\rbh{} spectral targets}} \\
		Bands & $\delta$ 0.5--4, $\theta$ 4--8, $\alpha$ 8--13, low-$\beta$ 13--20, high-$\beta$ 20--30, $\gamma$ 30--40\,Hz \\
		Spectral frontend & Hann STFT, $n_{\mathrm{perseg}}=n_{\mathrm{fft}}=200$, hop $=20$, frequencies $\leq40$\,Hz \\
		Target transform & $\log(1+x)$, normalized per channel \\
		Band-loss weight & $\lambda_{\mathrm{band}}=0.5$ \\
		\midrule
		\multicolumn{2}{@{}l}{\textbf{\method{} regularization}} \\
		Boundary interval & 1.0\,s / 200 samples \\
		RMS window & $q=32$ samples \\
		Penalty & Smooth-L1, $\beta=1$ \\
		Regularization weight & $\lambda_r=3.0$ \\
		Numerical constant & $\epsilon=10^{-6}$ \\
		\bottomrule
	\end{tabular}
\end{table*}

\section{Additional Downstream Transfer Protocols}

\label{app:downstream_variants}

Downstream evaluation follows the EEG-FM-Bench task definitions, subject splits, label mappings, and training pipeline~\citep{xiong2026eegfmbench}. The main paper reports \textsc{LP-Avg}; here we additionally evaluate (i) \textsc{LP-Flat}, where the frozen token grid is passed to a flatten-MLP head, (ii) \textsc{FT-Avg}, where the encoder is fine-tuned with average pooling, and (iii) \textsc{FT-Flat}, where the encoder is fine-tuned with the flatten-MLP head. All models use the same downstream settings within each protocol.

Models are trained for 30 epochs with AdamW (batch size 32, learning rate $2\times10^{-4}$, weight decay 0.01) and validation balanced accuracy is used for checkpoint selection.

We compare MANAS-2 with EEGPT~\citep{wang2024eegpt}, CSBrain~\citep{zhou2025csbrain}, CBraMod~\citep{wang2025cbramod}, BIOT~\citep{yang2023biot}, REVE~\citep{elouahidi2025reve}, and LaBraM~\citep{jiang2024lbm}.

\begin{table*}[!htbp]
\caption{ Frozen downstream transfer under \textsc{LP-Flat}. Values are mean test score (\%) at the best validation epoch, with variation across seeds.
}
\label{tab:frozen_downstream_transfer_flatten_mlp}
\centering
\renewcommand{\arraystretch}{0.90}
\setlength{\aboverulesep}{0.3ex}
\setlength{\belowrulesep}{0.45ex}
\scriptsize
\setlength{\tabcolsep}{2pt}
\resizebox{\textwidth}{!}{%
\begin{tabular}{@{}llccccccc@{}}
\toprule
Model & Metric & ADFTD & BCIC-2a & HMC & Motor$^{\ast}$ & Siena & Workload & MIMUL-11 \\
\midrule
EEGPT 
& Bal. Acc. & 40.1\scoreerr{2.0} & 38.1\scoreerr{1.7} & 60.5\scoreerr{1.6} & 56.9\scoreerr{0.9} & 81.0\scoreerr{4.9} & 63.3\scoreerr{1.1} & 45.2\scoreerr{0.7} \\
& F1/AUROC & 40.9\scoreerr{3.3} & 33.0\scoreerr{2.2} & 65.0\scoreerr{2.2} & 56.8\scoreerr{0.9} & \textbf{93.7}\scoreerr{1.7} & 68.8\scoreerr{1.2} & \textbf{54.4}\scoreerr{0.6} \\
& $\kappa$/AUC-PR & 14.0\scoreerr{3.7} & 17.4\scoreerr{2.2} & 55.4\scoreerr{3.0} & 42.5\scoreerr{1.1} & 69.7\scoreerr{5.0} & 46.9\scoreerr{1.9} & 22.3\scoreerr{0.6} \\
\midrule
CSBrain 
& Bal. Acc. & 51.0\scoreerr{0.8} & 28.2\scoreerr{3.3} & 64.8\scoreerr{1.0} & 26.7\scoreerr{0.8} & 50.0\scoreerr{0.0} & 52.1\scoreerr{0.8} & 34.9\scoreerr{1.0} \\
& F1/AUROC & \textbf{53.5}\scoreerr{0.7} & 15.6\scoreerr{4.9} & 67.0\scoreerr{1.7} & 19.0\scoreerr{2.8} & 48.1\scoreerr{18.3} & 54.2\scoreerr{1.4} & 39.8\scoreerr{1.8} \\
& $\kappa$/AUC-PR & 28.5\scoreerr{1.4} & 4.3\scoreerr{4.4} & 58.2\scoreerr{1.5} & 2.2\scoreerr{1.0} & 2.6\scoreerr{1.8} & 32.7\scoreerr{1.0} & 2.8\scoreerr{1.6} \\
\midrule
CBraMod 
& Bal. Acc. & 37.2\scoreerr{1.7} & 36.5\scoreerr{0.3} & 56.8\scoreerr{0.1} & 53.1\scoreerr{0.4} & 67.4\scoreerr{1.7} & 48.1\scoreerr{0.6} & 44.8\scoreerr{0.6} \\
& F1/AUROC & 40.6\scoreerr{2.1} & 34.5\scoreerr{0.8} & 61.4\scoreerr{0.2} & 52.7\scoreerr{0.5} & 89.3\scoreerr{1.9} & 49.1\scoreerr{0.2} & 51.1\scoreerr{0.7} \\
& $\kappa$/AUC-PR & 9.2\scoreerr{2.8} & 15.4\scoreerr{0.4} & 50.5\scoreerr{0.2} & 37.5\scoreerr{0.5} & 40.8\scoreerr{6.3} & 27.0\scoreerr{0.2} & 18.7\scoreerr{1.0} \\
\midrule
BIOT 
& Bal. Acc. & 43.7\scoreerr{2.5} & 26.3\scoreerr{1.4} & 64.7\scoreerr{0.4} & 27.6\scoreerr{1.2} & 65.6\scoreerr{3.5} & 51.4\scoreerr{2.8} & 37.5\scoreerr{1.9} \\
& F1/AUROC & 46.8\scoreerr{2.1} & 16.0\scoreerr{1.8} & 68.7\scoreerr{0.4} & 26.4\scoreerr{0.9} & 82.1\scoreerr{2.9} & 53.7\scoreerr{3.3} & 43.1\scoreerr{1.4} \\
& $\kappa$/AUC-PR & 18.1\scoreerr{3.5} & 1.7\scoreerr{1.9} & 59.7\scoreerr{0.4} & 3.4\scoreerr{1.6} & 39.2\scoreerr{5.7} & 34.1\scoreerr{4.8} & 6.2\scoreerr{3.2} \\
\midrule
REVE 
& Bal. Acc. & 37.9\scoreerr{7.8} & 33.9\scoreerr{1.1} & 60.2\scoreerr{1.1} & 47.6\scoreerr{1.6} & 63.4\scoreerr{3.1} & \textbf{64.4}\scoreerr{2.7} & 43.4\scoreerr{2.6} \\
& F1/AUROC & 37.9\scoreerr{9.8} & 25.8\scoreerr{1.8} & 65.8\scoreerr{0.5} & 46.9\scoreerr{2.4} & 67.0\scoreerr{3.9} & 66.2\scoreerr{0.8} & 47.1\scoreerr{3.7} \\
& $\kappa$/AUC-PR & 9.6\scoreerr{14.7} & 11.8\scoreerr{1.4} & 55.5\scoreerr{0.7} & 30.1\scoreerr{2.1} & 19.8\scoreerr{6.8} & \textbf{49.1}\scoreerr{3.4} & 16.0\scoreerr{3.3} \\
\midrule
LaBraM 
& Bal. Acc. & 35.3\scoreerr{1.1} & 25.4\scoreerr{0.6} & 37.9\scoreerr{1.1} & 25.0\scoreerr{0.4} & 50.0\scoreerr{0.0} & 48.9\scoreerr{1.7} & 37.6\scoreerr{0.4} \\
& F1/AUROC & 36.1\scoreerr{0.5} & 18.7\scoreerr{2.8} & 43.5\scoreerr{1.2} & 14.1\scoreerr{2.4} & 53.2\scoreerr{3.4} & 49.6\scoreerr{0.5} & 44.2\scoreerr{1.4} \\
& $\kappa$/AUC-PR & 4.8\scoreerr{1.4} & 0.5\scoreerr{0.8} & 28.1\scoreerr{1.4} & 0.0\scoreerr{0.5} & 1.4\scoreerr{0.3} & 27.8\scoreerr{0.8} & 8.4\scoreerr{0.7} \\
\midrule
\textbf{Ours: MANAS-2}
& Bal. Acc. & \textbf{51.9}\scoreerr{2.3} & \textbf{49.0}\scoreerr{2.0} & \textbf{72.3}\scoreerr{0.7} & \textbf{60.2}\scoreerr{1.3} & \textbf{86.5}\scoreerr{2.8} & 63.0\scoreerr{4.6} & \textbf{48.0}\scoreerr{0.5} \\
& F1/AUROC & 51.7\scoreerr{5.4} & \textbf{46.9}\scoreerr{2.8} & \textbf{74.8}\scoreerr{0.3} & \textbf{59.9}\scoreerr{1.4} & 92.8\scoreerr{2.1} & \textbf{70.3}\scoreerr{4.9} & 52.4\scoreerr{0.9} \\
& $\kappa$/AUC-PR & \textbf{31.5}\scoreerr{5.9} & \textbf{32.0}\scoreerr{2.6} & \textbf{66.9}\scoreerr{0.6} & \textbf{46.9}\scoreerr{1.8} & \textbf{74.6}\scoreerr{2.7} & 48.2\scoreerr{8.4} & \textbf{22.4}\scoreerr{0.9} \\
\bottomrule
\end{tabular}%
}
\vspace{2pt}
\begin{flushleft}
\scriptsize
$^{\ast}$Motor-MV was part of EEGPT's pretraining corpus; EEGPT's Motor results should therefore be interpreted with this overlap.
\end{flushleft}
\end{table*}
\begin{table*}[!htbp]
\caption{Full fine-tuning under \textsc{FT-Avg}. Values are mean test score (\%) at the best validation epoch, with variation across seeds.
}
\label{tab:full_ft_downstream_transfer_avg_pool}
\centering
\renewcommand{\arraystretch}{0.96}
\scriptsize
\setlength{\tabcolsep}{2pt}
\resizebox{\textwidth}{!}{%
\begin{tabular}{@{}llccccccc@{}}
\toprule
Model & Metric & ADFTD & BCIC-2a & HMC & Motor$^{\ast}$ & Siena & Workload & MIMUL-11 \\
\midrule
EEGPT 
& Bal. Acc. & 48.4\scoreerr{3.6} & 34.7\scoreerr{1.2} & 71.5\scoreerr{0.5} & 52.5\scoreerr{1.2} & 79.4\scoreerr{4.0} & 61.0\scoreerr{1.4} & 42.6\scoreerr{1.1} \\
& F1/AUROC & 51.0\scoreerr{2.6} & 27.5\scoreerr{1.9} & 73.0\scoreerr{0.5} & 51.9\scoreerr{1.3} & 91.6\scoreerr{3.3} & 69.8\scoreerr{2.5} & \textbf{52.1}\scoreerr{0.8} \\
& $\kappa$/AUC-PR & 26.7\scoreerr{3.9} & 13.0\scoreerr{1.7} & 65.6\scoreerr{0.5} & 36.6\scoreerr{1.6} & \textbf{68.7}\scoreerr{8.3} & \textbf{62.4}\scoreerr{1.8} & 18.3\scoreerr{0.8} \\
\midrule
CSBrain 
& Bal. Acc. & 40.6\scoreerr{6.2} & 35.1\scoreerr{1.4} & 70.9\scoreerr{0.5} & 42.5\scoreerr{4.8} & 70.2\scoreerr{4.1} & 65.5\scoreerr{3.8} & 39.0\scoreerr{1.6} \\
& F1/AUROC & 43.4\scoreerr{5.7} & 24.9\scoreerr{1.5} & 73.2\scoreerr{0.8} & 42.2\scoreerr{5.1} & 88.7\scoreerr{1.5} & 69.4\scoreerr{2.9} & 40.6\scoreerr{4.0} \\
& $\kappa$/AUC-PR & 12.8\scoreerr{10.0} & 13.5\scoreerr{1.8} & 65.3\scoreerr{0.8} & 23.4\scoreerr{6.4} & 56.6\scoreerr{6.8} & 59.3\scoreerr{5.9} & 7.9\scoreerr{2.7} \\
\midrule
CBraMod 
& Bal. Acc. & 30.3\scoreerr{3.0} & 25.3\scoreerr{0.3} & 53.0\scoreerr{1.0} & 31.4\scoreerr{1.6} & \textbf{84.9}\scoreerr{0.9} & 56.3\scoreerr{4.5} & 42.6\scoreerr{0.4} \\
& F1/AUROC & 26.0\scoreerr{4.2} & 10.9\scoreerr{0.6} & 52.2\scoreerr{2.0} & 28.5\scoreerr{1.9} & \textbf{93.5}\scoreerr{0.6} & 58.5\scoreerr{2.2} & 48.2\scoreerr{2.5} \\
& $\kappa$/AUC-PR & -5.7\scoreerr{5.6} & 0.4\scoreerr{0.4} & 42.5\scoreerr{2.0} & 8.5\scoreerr{2.1} & 42.7\scoreerr{8.3} & 36.8\scoreerr{1.1} & 14.8\scoreerr{1.4} \\
\midrule
BIOT 
& Bal. Acc. & 45.6\scoreerr{2.3} & 27.1\scoreerr{2.6} & 68.9\scoreerr{1.3} & 26.6\scoreerr{1.1} & 64.9\scoreerr{5.8} & 60.8\scoreerr{7.7} & 37.5\scoreerr{1.4} \\
& F1/AUROC & 48.2\scoreerr{2.3} & 18.2\scoreerr{1.8} & 72.5\scoreerr{1.2} & 24.9\scoreerr{1.8} & 83.6\scoreerr{5.9} & 65.4\scoreerr{10.2} & 41.6\scoreerr{1.2} \\
& $\kappa$/AUC-PR & 21.0\scoreerr{3.8} & 2.7\scoreerr{3.4} & 64.4\scoreerr{1.6} & 2.2\scoreerr{1.4} & 36.0\scoreerr{11.0} & 41.9\scoreerr{10.8} & 6.3\scoreerr{2.0} \\
\midrule
REVE 
& Bal. Acc. & 40.5\scoreerr{1.5} & 28.4\scoreerr{1.7} & 69.4\scoreerr{0.5} & 28.5\scoreerr{0.8} & 70.3\scoreerr{4.2} & 67.7\scoreerr{4.4} & 42.4\scoreerr{1.0} \\
& F1/AUROC & 43.8\scoreerr{1.2} & 17.5\scoreerr{2.8} & 71.8\scoreerr{1.0} & 28.0\scoreerr{0.7} & 84.0\scoreerr{6.2} & 73.8\scoreerr{3.8} & 45.8\scoreerr{4.2} \\
& $\kappa$/AUC-PR & 13.6\scoreerr{2.1} & 4.5\scoreerr{2.2} & 63.4\scoreerr{1.2} & 4.7\scoreerr{1.0} & 44.6\scoreerr{10.7} & 53.8\scoreerr{7.8} & 13.6\scoreerr{2.2} \\
\midrule
LaBraM 
& Bal. Acc. & 28.0\scoreerr{3.7} & 28.3\scoreerr{1.0} & 64.0\scoreerr{1.1} & 27.7\scoreerr{0.8} & 63.7\scoreerr{3.6} & 52.2\scoreerr{3.3} & 37.3\scoreerr{0.7} \\
& F1/AUROC & 28.6\scoreerr{4.2} & 26.3\scoreerr{1.6} & 66.0\scoreerr{2.0} & 23.1\scoreerr{3.2} & 88.5\scoreerr{2.5} & 54.7\scoreerr{3.1} & 43.4\scoreerr{1.1} \\
& $\kappa$/AUC-PR & -6.2\scoreerr{4.4} & 4.3\scoreerr{1.4} & 57.4\scoreerr{1.5} & 3.5\scoreerr{1.1} & 24.1\scoreerr{7.3} & 32.2\scoreerr{1.7} & 6.9\scoreerr{1.3} \\
\midrule
\textbf{Ours: MANAS-2} 
& Bal. Acc. & \textbf{50.8}\scoreerr{5.5} & \textbf{41.5}\scoreerr{2.6} & \textbf{73.5}\scoreerr{1.0} & \textbf{62.0}\scoreerr{0.5} & 81.9\scoreerr{4.1} & \textbf{70.3}\scoreerr{4.0} & \textbf{48.1}\scoreerr{0.5} \\
& F1/AUROC & \textbf{53.1}\scoreerr{5.2} & \textbf{35.3}\scoreerr{3.6} & \textbf{76.5}\scoreerr{0.5} & \textbf{61.9}\scoreerr{0.6} & 91.4\scoreerr{1.3} & \textbf{79.6}\scoreerr{2.2} & 49.7\scoreerr{2.9} \\
& $\kappa$/AUC-PR & \textbf{30.1}\scoreerr{8.3} & \textbf{22.0}\scoreerr{3.5} & \textbf{69.0}\scoreerr{0.6} & \textbf{49.3}\scoreerr{0.7} & 65.1\scoreerr{7.6} & 61.2\scoreerr{3.1} & \textbf{20.9}\scoreerr{1.5} \\
\bottomrule
\end{tabular}%
}
\vspace{2pt}
\begin{flushleft}
\scriptsize
$^{\ast}$Motor-MV was part of EEGPT's pretraining corpus; EEGPT's Motor results should therefore be interpreted with this overlap.
\end{flushleft}
\end{table*}
\begin{table*}[!htbp]
\caption{Full fine-tuning under \textsc{FT-Flat}. Values are mean test score (\%) at the best validation epoch, with variation across seeds.
}
\label{tab:full_ft_downstream_transfer_flatten_mlp}
\centering
\renewcommand{\arraystretch}{0.96}
\scriptsize
\setlength{\tabcolsep}{2pt}
\resizebox{\textwidth}{!}{%
\begin{tabular}{@{}llccccccc@{}}
\toprule
Model & Metric & ADFTD & BCIC-2a & HMC & Motor$^{\ast}$ & Siena & Workload & MIMUL-11 \\
\midrule
EEGPT 
& Bal. Acc. & 48.6\scoreerr{2.1} & 48.3\scoreerr{2.3} & 67.3\scoreerr{0.9} & 62.7\scoreerr{1.5} & 82.5\scoreerr{7.3} & 61.7\scoreerr{1.9} & 48.1\scoreerr{1.0} \\
& F1/AUROC & 52.1\scoreerr{2.3} & 45.5\scoreerr{2.7} & 70.3\scoreerr{0.6} & 62.4\scoreerr{1.6} & 89.9\scoreerr{5.7} & 66.2\scoreerr{1.0} & \textbf{57.8}\scoreerr{0.9} \\
& $\kappa$/AUC-PR & 26.4\scoreerr{4.0} & 31.0\scoreerr{3.0} & 61.6\scoreerr{1.0} & 50.3\scoreerr{2.0} & 69.6\scoreerr{10.0} & \textbf{53.3}\scoreerr{2.1} & \textbf{28.8}\scoreerr{1.2} \\
\midrule
CSBrain 
& Bal. Acc. & 45.3\scoreerr{2.2} & 47.0\scoreerr{1.2} & 70.0\scoreerr{0.7} & 61.4\scoreerr{0.6} & 63.0\scoreerr{2.9} & 56.8\scoreerr{4.9} & 39.0\scoreerr{1.2} \\
& F1/AUROC & 48.2\scoreerr{2.4} & 43.6\scoreerr{1.2} & 72.1\scoreerr{1.0} & 61.3\scoreerr{0.7} & 87.6\scoreerr{1.7} & 59.1\scoreerr{4.0} & 43.9\scoreerr{1.6} \\
& $\kappa$/AUC-PR & 19.8\scoreerr{3.6} & 29.4\scoreerr{1.6} & 64.2\scoreerr{0.8} & 48.5\scoreerr{0.8} & 40.9\scoreerr{6.7} & 40.3\scoreerr{3.9} & 9.3\scoreerr{1.7} \\
\midrule
CBraMod 
& Bal. Acc. & 41.1\scoreerr{3.7} & 43.5\scoreerr{0.7} & 68.1\scoreerr{0.9} & 58.8\scoreerr{0.4} & 82.2\scoreerr{2.6} & 55.4\scoreerr{2.1} & 44.7\scoreerr{0.4} \\
& F1/AUROC & 35.6\scoreerr{5.3} & 40.8\scoreerr{0.7} & 70.4\scoreerr{1.5} & 58.6\scoreerr{0.4} & \textbf{93.6}\scoreerr{4.4} & 62.6\scoreerr{1.3} & 49.4\scoreerr{0.9} \\
& $\kappa$/AUC-PR & 10.6\scoreerr{5.0} & 24.6\scoreerr{0.9} & 62.5\scoreerr{2.1} & 45.0\scoreerr{0.5} & \textbf{75.5}\scoreerr{2.3} & 45.9\scoreerr{2.4} & 18.2\scoreerr{0.4} \\
\midrule
BIOT 
& Bal. Acc. & 46.6\scoreerr{3.1} & 26.2\scoreerr{2.1} & 68.8\scoreerr{0.4} & 28.0\scoreerr{0.8} & 61.1\scoreerr{1.8} & 57.4\scoreerr{6.9} & 37.7\scoreerr{0.7} \\
& F1/AUROC & 49.5\scoreerr{2.4} & 16.1\scoreerr{1.8} & 72.4\scoreerr{0.6} & 25.8\scoreerr{0.7} & 83.7\scoreerr{1.6} & 62.0\scoreerr{8.7} & 41.9\scoreerr{2.1} \\
& $\kappa$/AUC-PR & 22.8\scoreerr{3.7} & 1.6\scoreerr{2.8} & 64.2\scoreerr{1.0} & 4.0\scoreerr{1.1} & 33.8\scoreerr{2.7} & 39.3\scoreerr{10.1} & 6.4\scoreerr{1.1} \\
\midrule
REVE 
& Bal. Acc. & 39.4\scoreerr{7.7} & 33.0\scoreerr{2.7} & 68.7\scoreerr{0.8} & 55.3\scoreerr{0.8} & 61.1\scoreerr{2.0} & 61.0\scoreerr{3.3} & 43.8\scoreerr{2.5} \\
& F1/AUROC & 37.4\scoreerr{10.8} & 24.1\scoreerr{3.9} & 72.1\scoreerr{0.8} & 55.5\scoreerr{0.7} & 64.7\scoreerr{1.5} & 65.0\scoreerr{3.6} & 45.3\scoreerr{3.9} \\
& $\kappa$/AUC-PR & 12.6\scoreerr{14.6} & 10.7\scoreerr{3.6} & 63.3\scoreerr{0.7} & 40.4\scoreerr{1.1} & 22.1\scoreerr{2.4} & 44.5\scoreerr{8.2} & 15.5\scoreerr{2.9} \\
\midrule
LaBraM 
& Bal. Acc. & 37.4\scoreerr{3.0} & 33.4\scoreerr{1.3} & 63.4\scoreerr{0.6} & 55.2\scoreerr{0.5} & 51.3\scoreerr{3.1} & 52.9\scoreerr{2.9} & 40.9\scoreerr{0.9} \\
& F1/AUROC & 41.1\scoreerr{2.9} & 33.1\scoreerr{1.4} & 66.0\scoreerr{1.3} & 55.1\scoreerr{0.5} & 60.3\scoreerr{12.9} & 54.0\scoreerr{3.8} & 50.0\scoreerr{0.8} \\
& $\kappa$/AUC-PR & 10.5\scoreerr{5.0} & 11.2\scoreerr{1.8} & 57.0\scoreerr{1.0} & 40.3\scoreerr{0.7} & 5.1\scoreerr{3.5} & 32.9\scoreerr{3.0} & 14.9\scoreerr{1.1} \\
\midrule
\textbf{Ours: MANAS-2} 
& Bal. Acc. & \textbf{52.1}\scoreerr{3.4} & \textbf{58.3}\scoreerr{3.3} & \textbf{73.8}\scoreerr{0.7} & \textbf{66.3}\scoreerr{0.7} & \textbf{83.4}\scoreerr{4.3} & \textbf{65.4}\scoreerr{3.6} & \textbf{50.3}\scoreerr{1.6} \\
& F1/AUROC & \textbf{54.3}\scoreerr{3.4} & \textbf{57.4}\scoreerr{3.9} & \textbf{76.4}\scoreerr{0.3} & \textbf{66.4}\scoreerr{0.6} & 91.1\scoreerr{2.5} & \textbf{72.1}\scoreerr{4.7} & 53.1\scoreerr{2.3} \\
& $\kappa$/AUC-PR & \textbf{35.8}\scoreerr{6.1} & \textbf{44.4}\scoreerr{4.4} & \textbf{68.9}\scoreerr{0.5} & \textbf{55.1}\scoreerr{0.9} & 72.6\scoreerr{3.7} & 48.9\scoreerr{9.9} & 25.1\scoreerr{2.7} \\
\bottomrule
\end{tabular}%
}
\vspace{2pt}
\begin{flushleft}
\scriptsize
$^{\ast}$Motor-MV was part of EEGPT's pretraining corpus; EEGPT's Motor results should therefore be interpreted with this overlap.
\end{flushleft}
\end{table*}
\FloatBarrier

Across these alternative probing and fine-tuning settings, \textsc{MANAS-2} remains competitive or strongest across most datasets, indicating that its downstream performance is not specific to the primary average-pooling protocol.

\FloatBarrier

\section{Objective-family and architecture ablations}
\label{app:objective_family}

We evaluate alternative ways of introducing spectral structure into masked EEG pretraining to isolate the contributions of the \rbh{} architecture and \method{}. Table~\ref{tab:model_family} summarizes the objective families considered.

\begin{table*}[!htbp]
\caption{Objective-family and architecture ablations.}
\label{tab:model_family}
\centering
\tiny
\setlength{\tabcolsep}{2pt}
\resizebox{0.95\textwidth}{!}{%
\begin{tabular}{@{}llll@{}}
\toprule
Model & Token space & Primary + secondary & Extra objective \\
\midrule
\textsc{Temporal} & Temporal patches & Temporal recon + global & --- \\
\textsc{Temporal+STFT} & Temporal patches & Temporal recon + global & Multi-resolution STFT magnitude+phase loss \\
\textsc{BandCube} & BandCube & Spectral recon + global & --- \\
\textsc{BandCube-T} & BandCube & Spectral recon + global & Temporal chunks \\
\textsc{RBH} & Temporal patches & Temporal recon + global & Band targets, dedicated decoder \\
\textsc{\method{}-Temporal} & Temporal patches & Temporal recon + global & Local RMS-energy regularizer \\
\textsc{MANAS-2} & Temporal patches & Temporal recon + global & Band targets + RMS-energy regularizer \\
\bottomrule
\end{tabular}%
}
\end{table*}

\subsection{Naive Temporal+STFT supervision}
\label{app:naive_stft}

\textsc{Temporal+STFT} tests whether spectral supervision can be added directly to a temporal MAE without introducing a dedicated spectral decoder. Masked temporal predictions receive an additional multi-resolution STFT magnitude-and-phase loss at resolutions $\{32,64,128\}$, with equal weighting of magnitude and phase terms; all other training settings follow the temporal baseline.

\begin{table*}[!htbp]
	\caption{
		Naive spectral-loss ablation. \textsc{Temporal+STFT} adds a multi-resolution STFT magnitude-and-phase loss to masked temporal predictions without introducing a dedicated spectral decoder. Mean test balanced accuracy (\%) is reported.
	}
	\label{tab:naive_stft}
	\centering
	\scriptsize
	\setlength{\tabcolsep}{4pt}
	\begin{tabular}{@{}lccccccc@{}}
		\toprule
		Model & ADFTD & BCIC-2a & HMC & Motor-MV & Siena & Workload & MIMUL-11 \\
		\midrule
		\textsc{Temporal-MAE} 
		& \textbf{49.2}\scoreerr{3.0} 
		& 28.5\scoreerr{1.0} 
		& \textbf{65.3}\scoreerr{0.2} 
		& \textbf{36.3}\scoreerr{0.2} 
		& \textbf{76.1}\scoreerr{2.0} 
		& \textbf{65.6}\scoreerr{0.5} 
		& \textbf{37.5}\scoreerr{0.9} \\
		\textsc{Temporal+STFT} 
		& 41.5\scoreerr{2.0} 
		& \textbf{29.3}\scoreerr{0.4} 
		& 59.7\scoreerr{0.3} 
		& 33.0\scoreerr{0.6} 
		& 51.0\scoreerr{1.1} 
		& 51.4\scoreerr{0.4} 
		& 35.6\scoreerr{1.0} \\
		\bottomrule
	\end{tabular}
\end{table*}

Table~\ref{tab:naive_stft} shows that direct STFT supervision underperforms the temporal MAE on six of seven datasets. This motivates the \rbh{} design: spectral supervision is more effective when provided through a dedicated spectral decoding pathway rather than added directly to temporal waveform reconstruction.

\subsection{BandCube spectral tokenization}
\label{app:bandcube}

\textsc{BandCube} tests a fully spectral alternative to temporal-patch tokenization. Its name reflects its central construction: each EEG window is represented as a three-dimensional \emph{channel $\times$ time $\times$ frequency-band} cube, and tokenization, masking, and reconstruction are defined over this joint 3D structure rather than over temporal waveform patches.

A frozen Hann-window STFT ($n_{\mathrm{fft}}=n_{\mathrm{perseg}}=200$, hop $=20$) is applied independently to each channel, restricted to frequencies $\leq40$\,Hz, log-transformed, normalized, and pooled into the six canonical EEG bands. For each sample this produces a spectral cube
\begin{equation}
	\mathbf{G}\in\mathbb{R}^{C\times M\times K},
	\qquad M=91,\quad K=6,
\end{equation}
where the three axes correspond to channel, spectral time, and frequency band.

The cube is partitioned into spectral macro-tokens spanning 12 STFT frames and one frequency band, with non-overlapping stride 12. After padding the temporal axis to 96 frames, this gives eight temporal groups, so each token is indexed by a location $(c,g,k)$ in the channel--time--band cube. Value embeddings are combined with channel--time and band positional encodings before being passed to the Transformer.

Importantly, the cube structure also defines the pretraining task: masking is sampled jointly over channel, time, and band locations, and the decoder reconstructs the masked spectral macro-tokens at their corresponding 3D positions using an L1 objective. The token grid is flattened only for Transformer processing; its channel--time--band organization is retained for masking, positional encoding, and reconstruction. As in the temporal MAE, a global auxiliary decoder additionally predicts masked targets.

Thus, \textsc{BandCube} represents the opposite architectural choice to \rbh{}: spectral structure is built directly into the encoder token space and reconstruction objective, rather than retaining temporal waveform tokens and supplying spectral information through a dedicated auxiliary pathway.

\subsection{BandCube temporal hybrid}
\label{app:bandcube_temporal}

\textsc{BandCube-T} retains the complete BandCube formulation: encoder tokens, masking, and primary reconstruction remain defined over the three-dimensional channel $\times$ time $\times$ frequency-band cube. It adds only an auxiliary temporal decoder to test whether an explicit waveform target can compensate for information lost by spectral tokenization.

For each channel--time group, the encoded cube is pooled across the six band positions to obtain one temporal representation. A separate decoder uses this representation to reconstruct the corresponding 2.1\,s raw-waveform segment. Temporal reconstruction is applied to groups with at least four masked band tokens; all other BandCube objectives remain unchanged.

\begin{table*}[!htbp]
	\caption{
		Spectral-tokenization ablation under full fine-tuning with average pooling. \textsc{BandCube} uses channel $\times$ time $\times$ band tokens, while \textsc{BandCube-T} additionally reconstructs temporal waveform segments. Mean test balanced accuracy (\%) is reported.
	}
	\label{tab:bandcube_ftavg}
	\centering
	\scriptsize
	\setlength{\tabcolsep}{4pt}
	\begin{tabular}{@{}lccccccc@{}}
		\toprule
		Model & ADFTD & BCIC-2a & Motor-MV & Siena & Workload & MIMUL-11 & Avg. \\
		\midrule
		\textsc{Temporal-MAE} 
		& 57.4\scoreerr{3.5} 
		& \textbf{47.8}\scoreerr{2.9} 
		& \textbf{62.7}\scoreerr{1.3} 
		& 82.5\scoreerr{5.1} 
		& \textbf{66.4}\scoreerr{2.4} 
		& \textbf{48.1}\scoreerr{2.1} 
		& \textbf{60.8}\scoreerr{2.9} \\
		
		\textsc{BandCube} 
		& \textbf{62.9}\scoreerr{4.0} 
		& 37.9\scoreerr{4.5} 
		& 49.4\scoreerr{0.7} 
		& \textbf{84.0}\scoreerr{3.1} 
		& 65.5\scoreerr{5.7} 
		& 41.4\scoreerr{1.6} 
		& 56.9\scoreerr{3.3} \\
		
		\textsc{BandCube-T} 
		& 56.7\scoreerr{3.1} 
		& 34.8\scoreerr{3.6} 
		& 47.8\scoreerr{1.1} 
		& 82.7\scoreerr{4.6} 
		& 63.4\scoreerr{2.6} 
		& 38.9\scoreerr{2.5} 
		& 54.1\scoreerr{2.9} \\
		\bottomrule
	\end{tabular}
\end{table*}

Spectral tokenization is competitive on some datasets, but reduces mean balanced accuracy from 60.8\% for \textsc{Temporal-MAE} to 56.9\% for \textsc{BandCube}, with particularly large losses on BCIC-2a and Motor-MV. Adding temporal reconstruction does not recover this loss: \textsc{BandCube-T} reaches 54.1\% average balanced accuracy. This is consistent with motor-imagery EEG being expressed through time-varying event-related desynchronization/synchronization (ERD/ERS) of sensorimotor rhythms~\citep{pfurtscheller1999event}, which may be less faithfully preserved when the encoder operates on pooled spectral macro-tokens. These results motivate \rbh{} to retain temporal waveform tokens in the encoder while introducing spectral supervision through a dedicated decoder.

\subsection{From temporal reconstruction to RBH to MANAS-2}
\label{app:rbh_conrec_lpavg}

This ablation isolates the two components of \textsc{MANAS-2}. \textsc{Temporal-MAE} provides the temporal reconstruction baseline, \method{}-Temporal adds the local RMS-energy regularizer without spectral supervision, \rbh{} adds a dedicated spectral decoder without \method{}, and \textsc{MANAS-2} combines both.

\begin{table*}[!htbp]
	\caption{
		Frozen \textsc{LP-Avg} ablation from temporal reconstruction to \rbh{} and \textsc{MANAS-2}. Values are mean test score (\%) at the best validation epoch, with variation across seeds. Bold marks the best mean within this four-model comparison.
	}
	\label{tab:rbh_conrec_lpavg}
	\centering
	\scriptsize
	\setlength{\tabcolsep}{2pt}
	\resizebox{\textwidth}{!}{%
		\begin{tabular}{@{}llcccccccc@{}}
			\toprule
			Model & Metric & ADFTD & BCIC-2a & HMC & Motor-MV & Siena & Workload & MIMUL-11 & Avg. \\
			\midrule
			
			\textsc{Temporal-MAE}
			& Bal. Acc. & \textbf{49.2}\scoreerr{3.0} & 28.5\scoreerr{1.0} & 65.3\scoreerr{0.2} & 36.3\scoreerr{0.2} & 76.1\scoreerr{2.0} & 65.6\scoreerr{0.5} & 37.5\scoreerr{0.9} & 51.2\scoreerr{0.6} \\
			& F1/AUROC & \textbf{51.1}\scoreerr{4.1} & 19.6\scoreerr{1.9} & 69.6\scoreerr{0.5} & 36.2\scoreerr{0.2} & 93.5\scoreerr{0.5} & 73.8\scoreerr{0.4} & 44.5\scoreerr{1.6} & 55.5\scoreerr{0.7} \\
			& $\kappa$/AUC-PR & 29.3\scoreerr{4.8} & 4.6\scoreerr{1.3} & 61.2\scoreerr{0.4} & 15.0\scoreerr{0.3} & 62.7\scoreerr{3.3} & 41.5\scoreerr{0.5} & 7.1\scoreerr{1.8} & 31.6\scoreerr{0.9} \\
			\midrule
			
			\textsc{\method{}-Temporal}
			& Bal. Acc. & 48.0\scoreerr{2.1} & \textbf{31.1}\scoreerr{0.5} & 67.9\scoreerr{0.6} & \textbf{38.7}\scoreerr{0.3} & 79.2\scoreerr{4.3} & 70.0\scoreerr{1.3} & 37.3\scoreerr{0.8} & 53.2\scoreerr{0.7} \\
			& F1/AUROC & 49.3\scoreerr{2.0} & \textbf{23.7}\scoreerr{0.9} & 72.0\scoreerr{0.8} & \textbf{38.5}\scoreerr{0.4} & \textbf{94.4}\scoreerr{0.2} & 75.8\scoreerr{0.2} & 45.0\scoreerr{1.5} & 57.0\scoreerr{0.4} \\
			& $\kappa$/AUC-PR & 28.1\scoreerr{3.9} & \textbf{8.1}\scoreerr{0.7} & 64.1\scoreerr{1.0} & \textbf{18.3}\scoreerr{0.3} & 75.1\scoreerr{0.6} & 43.0\scoreerr{0.4} & 7.9\scoreerr{1.8} & 34.9\scoreerr{0.6} \\
			\midrule
			
			\textsc{RBH}
			& Bal. Acc. & 49.0\scoreerr{1.3} & 28.1\scoreerr{0.9} & 69.1\scoreerr{0.1} & 37.1\scoreerr{0.8} & \textbf{82.7}\scoreerr{1.6} & 69.9\scoreerr{2.8} & \textbf{39.8}\scoreerr{0.7} & 53.7\scoreerr{0.5} \\
			& F1/AUROC & 50.4\scoreerr{1.6} & 17.4\scoreerr{1.8} & 73.2\scoreerr{0.2} & 36.8\scoreerr{0.9} & 91.3\scoreerr{0.2} & 79.0\scoreerr{1.5} & 45.9\scoreerr{0.3} & 56.3\scoreerr{0.4} \\
			& $\kappa$/AUC-PR & \textbf{30.0}\scoreerr{2.4} & 4.1\scoreerr{1.2} & 65.5\scoreerr{0.4} & 16.1\scoreerr{1.1} & \textbf{75.9}\scoreerr{0.4} & 48.1\scoreerr{2.9} & \textbf{10.7}\scoreerr{0.6} & 35.8\scoreerr{0.6} \\
			\midrule
			
			\textbf{Ours: MANAS-2}
			& Bal. Acc. & 47.7\scoreerr{2.6} & 29.2\scoreerr{0.5} & \textbf{70.3}\scoreerr{0.3} & 37.7\scoreerr{0.7} & 80.0\scoreerr{1.2} & \textbf{77.8}\scoreerr{0.8} & 39.1\scoreerr{0.4} & \textbf{54.5}\scoreerr{0.4} \\
			& F1/AUROC & 49.3\scoreerr{3.9} & 20.5\scoreerr{0.9} & \textbf{74.5}\scoreerr{0.4} & 37.2\scoreerr{0.8} & 92.9\scoreerr{0.1} & \textbf{85.0}\scoreerr{0.7} & \textbf{46.1}\scoreerr{0.4} & \textbf{57.9}\scoreerr{0.6} \\
			& $\kappa$/AUC-PR & 27.3\scoreerr{4.8} & 5.6\scoreerr{0.7} & \textbf{66.9}\scoreerr{0.2} & 17.0\scoreerr{0.9} & 73.4\scoreerr{0.5} & \textbf{58.8}\scoreerr{2.0} & 9.9\scoreerr{0.5} & \textbf{37.0}\scoreerr{0.8} \\
			\bottomrule
		\end{tabular}%
	}
\end{table*}

Both components improve the temporal baseline on average. \method{}-Temporal raises mean balanced accuracy from 51.2\% to 53.2\%, while \rbh{} reaches 53.7\%. Combining the \rbh{} architecture with \method{} in \textsc{MANAS-2} gives the strongest average performance: 54.5\% balanced accuracy, 57.9\% F1/AUROC, and 37.0\% $\kappa$/AUC-PR. This supports complementary contributions from explicit spectral supervision and the reconstruction-space RMS-energy regularizer.

\FloatBarrier

\FloatBarrier
\section{Spectral Recoverability - Additional Details}
\label{app:stftbandflow_details}
\subsection{Probe Methodology and Dataset Details}
\paragraph{Probe methodology} For each frozen model, we extracted local latent tokens $z_{bpc}\in\mathbb{R}^D$ on the model's native patch-channel grid, with feature tensor $Z\in\mathbb{R}^{B\times P\times C\times D}$. Each $(b,p,c)$ token was treated as one probe example. For the STFT 6-band target, temporal EEG was converted to log-magnitude STFT frames using a 200-sample Hann window, hop 20, $n_{\mathrm{fft}}=200$, one-sided spectrum, $f_s=200$ Hz, and frequencies $\leq 40$ Hz. STFT frames were aligned to each model's latent patch grid by overlap-weighted temporal averaging: each latent patch received a weighted average of STFT frames, with weights proportional to the temporal overlap between the STFT frame and the latent patch interval. The aligned STFT representation was then averaged within six frequency bands: $\delta=[0.5,4)$, $\theta=[4,8)$, $\alpha=[8,13)$, low-$\beta=[13,20)$, high-$\beta=[20,30)$, and $\gamma=[30,40)$ Hz, giving $y^{\mathrm{band}}_{bpc}\in\mathbb{R}^6$. For bandflow, the target was the adjacent patch difference $y^{\mathrm{flow}}_{bpc}=y^{\mathrm{band}}_{b,p+1,c} -y^{\mathrm{band}}_{bpc}\in\mathbb{R}^6$; the corresponding input was the current local token $z_{bpc}$, with the feature tensor truncated to the first $P-1$ patch positions.

For the PAF, periodic component, and aperiodic exponent probes, we used 4-patch windows on the same patch grid.
For a window beginning at patch $p$, the probe input was the mean-pooled latent
$\bar z_{bpc}=K^{-1}\sum_{j=0}^{K-1}z_{b,p+j,c}$ with $K=4$, and the target was
computed from the corresponding raw EEG span. \footnote{Using these 4 second windows is why recovery for BCI-IV-2a is low in absolute terms (Table \ref{tab:latent_probes}). Since the BCI-IV-2a data is segmented in 4s windows, each example provides only a single short motor-imagery interval rather than multiple independent temporal windows over which spectral estimates can be averaged. Consequently, targets such as PAF and periodic spectral components are estimated from fewer samples and are intrinsically noisier, making them harder to recover from frozen latents even when relative differences between models remain consistent.} Let
$S_{bpc}(f)$ denote the Welch power spectral density estimated on that span.
The peak-alpha-frequency target was the location of the maximum PSD value in
the alpha band,
\[
y^{\mathrm{PAF}}_{bpc}
=\arg\max_{f\in[8,13]\ \mathrm{Hz}} S_{bpc}(f)
\in\mathbb{R}.
\]
For the aperiodic and periodic targets, we used the fixed-aperiodic FOOOF
decomposition of the PSD \citep{donoghue2020fooof}. In this parameterization, the log spectrum is
approximated as
\[
\log_{10}S_{bpc}(f)\approx
\underbrace{a_{bpc}-\chi_{bpc}\log_{10}f}_{A_{bpc}(f)}
+
\underbrace{\sum_m h_{bpcm}
	\exp\!\left[-\frac{(f-\mu_{bpcm})^2}{2\sigma_{bpcm}^2}\right]}_{G_{bpc}(f)} ,
\]
where $A_{bpc}(f)$ is the aperiodic background and $G_{bpc}(f)$ is the
periodic component. The aperiodic-exponent target was the scalar
$y^{\mathrm{aper}}_{bpc}=\chi_{bpc}$, while the periodic-component target was
the Gaussian component evaluated on a fixed frequency grid
$g_q\in\{1.0,1.5,\ldots,45.0\}$ Hz,
\[
y^{\mathrm{per}}_{bpc}
=\bigl(G_{bpc}(g_1),\ldots,G_{bpc}(g_{89})\bigr)\in\mathbb{R}^{89}.
\]
For linear probes, we flattened valid examples as
$X\in\mathbb{R}^{N\times D}$ and $Y\in\mathbb{R}^{N\times d_y}$, standardized
both $X$ and $Y$ using training-set mean and standard deviation, and fit a
multi-output ridge map $W_\lambda$ by
\[
W_\lambda=(\tilde X_{\mathrm{train}}^\top \tilde X_{\mathrm{train}}
+\lambda I)^{-1}\tilde X_{\mathrm{train}}^\top\tilde Y_{\mathrm{train}}.
\]
We selected $\lambda\in\{0.1,1,10,100\}$ by validation mean $R^2$,
inverse-transformed predictions to the original target scale, and reported test
$R^2$ averaged over target dimensions.

\paragraph{Datasets} All latent-space analyses are reported by averaging over all 7 downstream datasets used in the paper (Appendix \ref{app:downstream_datasets}), to ensure evaluation over a broad spectrum of unseen data. To give roughly equal weightage to all datasets, we take a subset of subjects and samples from each. For each dataset and seed, we sample up to 480 training windows, 192 validation windows, and 192 test windows, with subject-balanced sampling within each split. The same sampled split is used for every model and every probe target, so model comparisons are paired on identical data. We repeat this process over five data seeds, $s \in \{42,43,44,45,46\}$; each data seed corresponds to a different subject/sample subset drawn under the same caps. For MLP probes, we pair each data seed with a corresponding optimization seed, $s_{\mathrm{MLP}}\in\{1042,1043,1044,1045,1046\}$, which controls the MLP initialization and minibatch order. 

An expanded version of Table \ref{tab:latent_probes} with per-dataset results is shown in Table \ref{tab:latent_probes_expanded}.

\begin{table*}[!htbp]
	\caption{
		Ridge probe $R^2$ for temporal MAE, RBH, and MANAS-2 across spectral and temporal targets. Cells report mean $R^2 \pm$ 95\% confidence intervals. Dataset aliases: BC: BCI-IV-2a, MV: Motor-MV, WL: Workload, SI: Siena Scalp, HM: HMC, AD: ADFTD, MI: MIMUL11.
	}
	\label{tab:latent_probes_expanded}
	\centering
	\scriptsize
	\setlength{\tabcolsep}{2pt}
	\resizebox{\textwidth}{!}{%
		\begin{tabular}{@{}llcccccccc@{}}
			\toprule
			Metric & Model & BC & MV & WL & SI & HM & AD & MI & Avg. \\
			\midrule
			6-band STFT & Temporal MAE & .659\scoreerr{.02} & .773\scoreerr{.03} & .648\scoreerr{.01} & .818\scoreerr{.01} & .881\scoreerr{.03} & .660\scoreerr{.17} & .785\scoreerr{.01} & .747\scoreerr{.03} \\
			& RBH & .795\scoreerr{.01} & .871\scoreerr{.01} & .815\scoreerr{.00} & .900\scoreerr{.00} & .930\scoreerr{.02} & .829\scoreerr{.07} & .880\scoreerr{.00} & .860\scoreerr{.01} \\
			& MANAS-2 & \textbf{.883}\scoreerr{.00} & \textbf{.916}\scoreerr{.01} & \textbf{.892}\scoreerr{.00} & \textbf{.931}\scoreerr{.00} & \textbf{.953}\scoreerr{.01} & \textbf{.843}\scoreerr{.09} & \textbf{.927}\scoreerr{.00} & \textbf{.906}\scoreerr{.01} \\
			\midrule
			Bandflow STFT & Temporal MAE & .146\scoreerr{.01} & .188\scoreerr{.03} & .164\scoreerr{.01} & .177\scoreerr{.02} & .180\scoreerr{.03} & .118\scoreerr{.02} & .179\scoreerr{.01} & .165\scoreerr{.01} \\
			& RBH & .256\scoreerr{.02} & .309\scoreerr{.03} & .305\scoreerr{.01} & .269\scoreerr{.01} & .261\scoreerr{.03} & .264\scoreerr{.03} & .314\scoreerr{.00} & .283\scoreerr{.01} \\
			& MANAS-2 & \textbf{.334}\scoreerr{.02} & \textbf{.362}\scoreerr{.03} & \textbf{.369}\scoreerr{.01} & \textbf{.344}\scoreerr{.02} & \textbf{.363}\scoreerr{.03} & \textbf{.321}\scoreerr{.04} & \textbf{.382}\scoreerr{.00} & \textbf{.354}\scoreerr{.01} \\
			\midrule
			PAF & Temporal MAE & -.244\scoreerr{.11} & .009\scoreerr{.10} & .112\scoreerr{.07} & .038\scoreerr{.02} & .001\scoreerr{.15} & .103\scoreerr{.08} & .085\scoreerr{.03} & .015\scoreerr{.03} \\
			& RBH & -.204\scoreerr{.16} & .224\scoreerr{.05} & .280\scoreerr{.03} & .147\scoreerr{.02} & .120\scoreerr{.13} & .172\scoreerr{.05} & .206\scoreerr{.01} & .135\scoreerr{.03} \\
			& MANAS-2 & \textbf{-.088}\scoreerr{.08} & \textbf{.286}\scoreerr{.03} & \textbf{.358}\scoreerr{.02} & \textbf{.224}\scoreerr{.02} & \textbf{.210}\scoreerr{.11} & \textbf{.211}\scoreerr{.08} & \textbf{.257}\scoreerr{.02} & \textbf{.208}\scoreerr{.02} \\
			\midrule
			Periodic component & Temporal MAE & -.248\scoreerr{.11} & .162\scoreerr{.05} & .151\scoreerr{.02} & .062\scoreerr{.02} & .166\scoreerr{.06} & .011\scoreerr{.15} & .164\scoreerr{.03} & .067\scoreerr{.04} \\
			& RBH & -.160\scoreerr{.10} & .315\scoreerr{.05} & .287\scoreerr{.01} & .239\scoreerr{.01} & .278\scoreerr{.07} & .169\scoreerr{.09} & .295\scoreerr{.01} & .203\scoreerr{.03} \\
			& MANAS-2 & \textbf{-.092}\scoreerr{.09} & \textbf{.358}\scoreerr{.04} & \textbf{.333}\scoreerr{.01} & \textbf{.297}\scoreerr{.02} & \textbf{.335}\scoreerr{.07} & \textbf{.208}\scoreerr{.06} & \textbf{.337}\scoreerr{.01} & \textbf{.254}\scoreerr{.02} \\
			\midrule
			Aperiodic exponent & Temporal MAE & .881\scoreerr{.01} & .744\scoreerr{.05} & .598\scoreerr{.03} & .710\scoreerr{.02} & .831\scoreerr{.05} & .261\scoreerr{.81} & .720\scoreerr{.02} & .678\scoreerr{.11} \\
			& RBH & .902\scoreerr{.02} & .805\scoreerr{.04} & .718\scoreerr{.01} & .780\scoreerr{.02} & .879\scoreerr{.03} & .665\scoreerr{.36} & .769\scoreerr{.01} & .788\scoreerr{.05} \\
			& MANAS-2 & .907\scoreerr{.03} & .818\scoreerr{.04} & .725\scoreerr{.02} & .790\scoreerr{.01} & .880\scoreerr{.04} & .655\scoreerr{.35} & .766\scoreerr{.01} & .792\scoreerr{.05} \\
			\midrule
			Temporal waveform & Temporal MAE & \textbf{.964}\scoreerr{.00} & \textbf{.908}\scoreerr{.02} & \textbf{.971}\scoreerr{.00} & \textbf{.881}\scoreerr{.00} & \textbf{.929}\scoreerr{.01} & \textbf{.941}\scoreerr{.02} & \textbf{.896}\scoreerr{.00} & \textbf{.927}\scoreerr{.00} \\
			& RBH & .959\scoreerr{.00} & .903\scoreerr{.02} & .970\scoreerr{.00} & .874\scoreerr{.00} & .917\scoreerr{.01} & .928\scoreerr{.02} & .892\scoreerr{.00} & .920\scoreerr{.00} \\
			& MANAS-2 & .958\scoreerr{.00} & .900\scoreerr{.02} & .968\scoreerr{.00} & .871\scoreerr{.00} & .910\scoreerr{.01} & .925\scoreerr{.02} & .890\scoreerr{.00} & .917\scoreerr{.00} \\
			\bottomrule
		\end{tabular}
	}
\end{table*}

\subsection{Additional methods}
We show the spectral recoverability effect induced by \method{} holds on multiple objectives and probes.
\paragraph{MLP probe} Here we show that the spectral features can be better recovered \textit{non-linearly} (as opposed to linearly through the ridge regression probes) as well when using \method{}. With standardized frozen latents as the input, we train a fixed two-layer MLP probe with one 256-unit GELU hidden layer (no dropout), trained with AdamW using learning rate \(10^{-3}\) and weight decay \(10^{-4}\). The probe was selected by validation MSE with patience 3 over at most 20 epochs; no probe hyperparameters were tuned per model or dataset.
\begin{table}[H]
	\caption{
		MLP probe $R^2$ for RBH and \textsc{MANAS-2} on the same spectral targets. Cells report mean $R^2$ across paired splits with
		variation shown next to each mean.
	}
	\label{tab:rbh_mlp_probes_expanded}
	\centering
	\scriptsize
	\setlength{\tabcolsep}{2pt}
	\resizebox{\textwidth}{!}{%
		\begin{tabular}{@{}llcccccccc@{}}
			\toprule
			Metric & Model & BC & MV & WL & SI & HM & AD & MI & Avg. \\
			\midrule
			6-band STFT & RBH & .766\scoreerr{.006} & .890\scoreerr{.011} & .803\scoreerr{.003} & .918\scoreerr{.003} & .910\scoreerr{.014} & .849\scoreerr{.045} & .894\scoreerr{.003} & .862\scoreerr{.009} \\
			& MANAS-2 & \textbf{.860}\scoreerr{.007} & \textbf{.941}\scoreerr{.006} & \textbf{.891}\scoreerr{.002} & \textbf{.953}\scoreerr{.001} & \textbf{.933}\scoreerr{.006} & \textbf{.892}\scoreerr{.048} & \textbf{.945}\scoreerr{.002} & \textbf{.916}\scoreerr{.008} \\
			\midrule
			Bandflow STFT & RBH & .226\scoreerr{.007} & .326\scoreerr{.026} & .293\scoreerr{.013} & .301\scoreerr{.018} & .261\scoreerr{.028} & .272\scoreerr{.030} & .315\scoreerr{.004} & .285\scoreerr{.006} \\
			& MANAS-2 & \textbf{.307}\scoreerr{.015} & \textbf{.406}\scoreerr{.025} & \textbf{.378}\scoreerr{.012} & \textbf{.395}\scoreerr{.022} & \textbf{.369}\scoreerr{.028} & \textbf{.344}\scoreerr{.031} & \textbf{.402}\scoreerr{.004} & \textbf{.372}\scoreerr{.005} \\
			\midrule
			PAF & RBH & .045\scoreerr{.049} & .246\scoreerr{.021} & .262\scoreerr{.065} & .140\scoreerr{.015} & .089\scoreerr{.112} & .181\scoreerr{.064} & .206\scoreerr{.013} & .167\scoreerr{.049} \\
			& MANAS-2& \textbf{.135}\scoreerr{.057} & \textbf{.309}\scoreerr{.029} & \textbf{.393}\scoreerr{.021} & \textbf{.227}\scoreerr{.014} & \textbf{.210}\scoreerr{.058} & \textbf{.224}\scoreerr{.076} & \textbf{.263}\scoreerr{.016} & \textbf{.252}\scoreerr{.039} \\
			\midrule
			Periodic component & RBH & .012\scoreerr{.050} & .333\scoreerr{.037} & .283\scoreerr{.009} & .270\scoreerr{.010} & .215\scoreerr{.052} & .181\scoreerr{.077} & .310\scoreerr{.011} & .229\scoreerr{.035} \\
			& MANAS-2 & \textbf{.063}\scoreerr{.033} & \textbf{.371}\scoreerr{.030} & \textbf{.323}\scoreerr{.012} & \textbf{.325}\scoreerr{.014} & \textbf{.242}\scoreerr{.046} & \textbf{.215}\scoreerr{.065} & \textbf{.352}\scoreerr{.008} & \textbf{.270}\scoreerr{.030} \\
			\midrule
			Aperiodic exponent & RBH & .837\scoreerr{.010} & .815\scoreerr{.027} & .680\scoreerr{.022} & .800\scoreerr{.011} & .776\scoreerr{.024} & .655\scoreerr{.191} & .789\scoreerr{.014} & .765\scoreerr{.043} \\
			& MANAS-2 & .810\scoreerr{.053} & .824\scoreerr{.019} & .675\scoreerr{.035} & .802\scoreerr{.007} & .795\scoreerr{.051} & .690\scoreerr{.131} & .798\scoreerr{.012} & .771\scoreerr{.044} \\
			\bottomrule
		\end{tabular}%
	}
\end{table}

As shown in Table \ref{tab:rbh_mlp_probes_expanded}, we see the \method{} effect holds even when allowing for non-linear recovery. Additionally, to assess statistical significance of the results for both the ridge and MLP probes, we computed paired dataset-level differences in $R^2$ between \textsc{MANAS-2} and RBH and tested whether the mean difference across datasets was nonzero using a two-sided one-sample $t$ test; all resulting row-level $p$ values were below $0.05$ except for the aperiodic exponent target.

\FloatBarrier
\section{Additional Latent Geometry Details}
\label{app:geometry_full}

This appendix provides additional details and visualizations for the synthetic latent-geometry analysis in Section~\ref{sec:synthetic_geometry}.

\paragraph{Synthetic inputs and latent pooling.}
We generate sinusoids spanning 0.5--40\,Hz in 0.5\,Hz increments, with $K=5$ repetitions per frequency. For the 18-channel condition, channel amplitudes are sampled with small Gaussian variation and phase is independently randomized across channels. We additionally evaluate a single-channel Cz condition to remove spatial effects. Because frequency remains constant within each input, this diagnostic isolates frequency-dependent latent organization from evolving spectral dynamics.

For model $m$, frequency $f$, and repetition $k$, the frozen encoder returns token latents
$E_m(\tilde{x}_{f,k},p)\in\mathbb{R}^{T_m\times C\times d}$. We average over latent time and channels to obtain one pooled representation per repetition, and then average across repetitions to obtain a centroid for each frequency:
\begin{equation}
	\mathbf{z}^{(m)}_{f,k}
	=
	\frac{1}{T_m C}
	\sum_{t=1}^{T_m}\sum_{c=1}^{C}
	E_m(\tilde{x}_{f,k},p)_{t,c,:},
	\qquad
	\bar{\mathbf{z}}^{(m)}_f
	=
	\frac{1}{K}\sum_{k=1}^{K}\mathbf{z}^{(m)}_{f,k}.
\end{equation}
The single-channel condition has lower spatial resolution than the data observed during pretraining and is therefore treated as a complementary diagnostic rather than an in-distribution input condition.

\begin{figure}[!htbp]
	\centering
	\includegraphics[width=\linewidth]{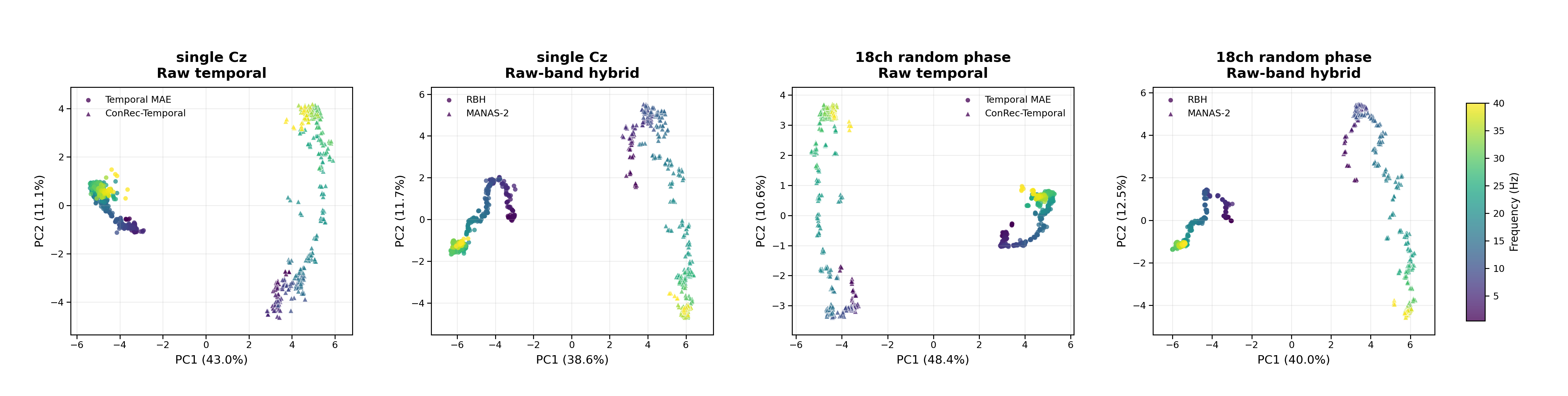}
	\caption{
		Shared-basis PCA of pooled latent representations for single-channel Cz and 18-channel random-phase synthetic sinusoids. Applying \method{} produces more extended and more clearly frequency-dependent latent trajectories in both the temporal-only and \rbh{} model families.
	}
	\label{fig:sinusoidpca}
\end{figure}

Figure~\ref{fig:sinusoidpca} provides a qualitative view of the latent trajectories induced by changing sinusoid frequency. In both architectural families, the \method{} variants occupy a larger and more visibly ordered trajectory, consistent with the stronger frequency-distance ordering reported in the main paper.

\begin{figure}[!htbp]
	\centering
	\includegraphics[width=\linewidth]{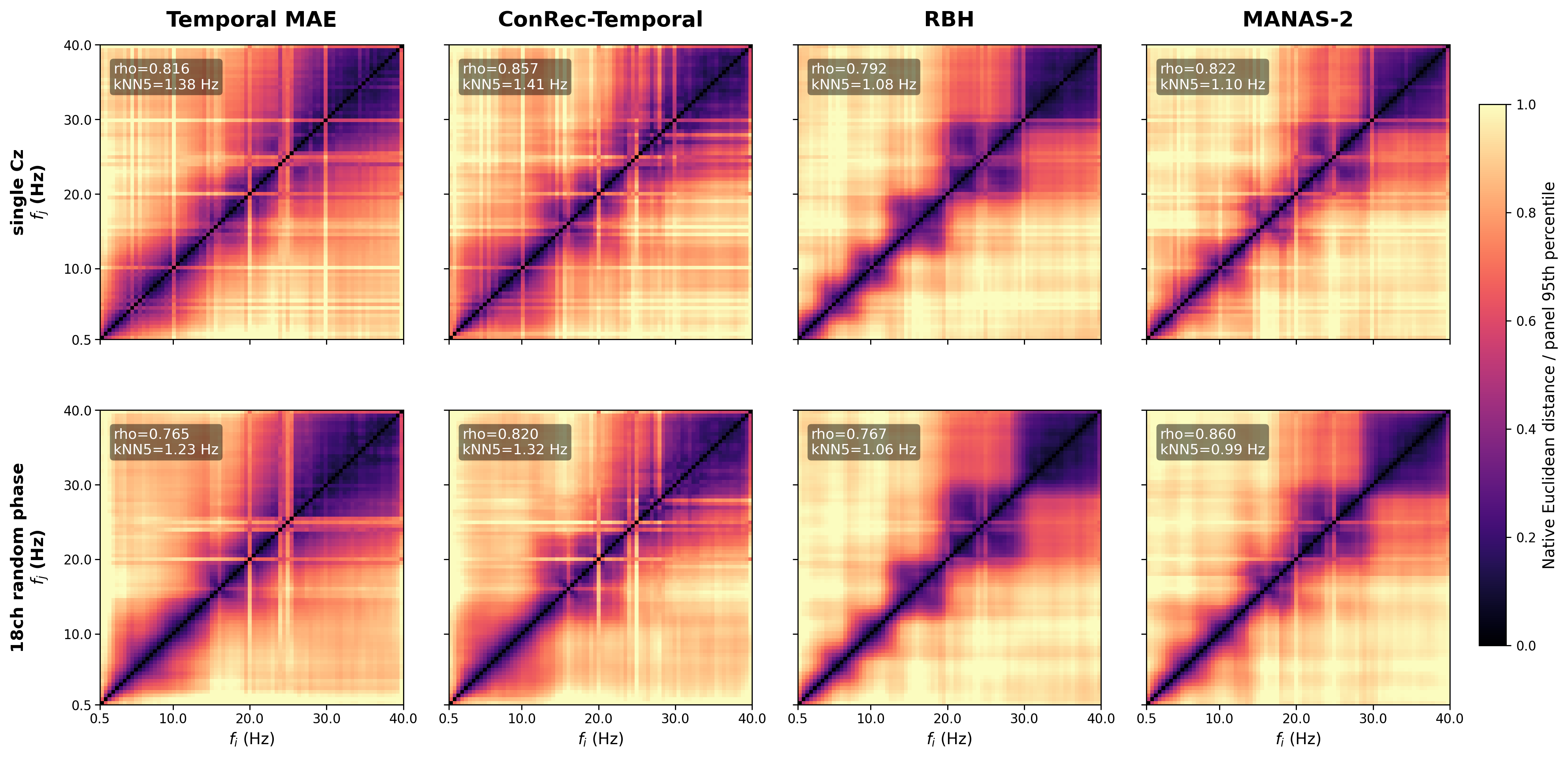}
	\caption{
		Pairwise Euclidean distances between frequency centroids in the latent space. More regular structure away from the diagonal indicates that latent distance tracks physical frequency separation more consistently.
	}
	\label{fig:euclid_heatmap}
\end{figure}

Figure~\ref{fig:euclid_heatmap} visualizes the same frequency-distance structure from a complementary perspective. The \method{} variants show a more regular progression of pairwise distances across frequencies, matching the improved Spearman $\rho_f$ values reported in Section~\ref{sec:synthetic_geometry}. Together, the PCA and heatmap views support the conclusion that \method{} reshapes latent spectral organization rather than merely changing overall spread.
\FloatBarrier

\end{document}